\documentclass[lettersize,journal]{IEEEtran}

\usepackage{url}
\usepackage{graphicx}
\usepackage{amsmath,amssymb}
\usepackage{multirow}
\usepackage{bbm} 
\usepackage{tabularx}
\usepackage{booktabs}
\usepackage{textcomp}
\usepackage{stfloats}
\usepackage{url}
\usepackage{verbatim}
\usepackage{graphicx}
\usepackage{cite}
\usepackage{subfig}

\usepackage{pifont}
\newcommand{\cmark}{\ding{51}}%
\usepackage[T1]{fontenc}

\usepackage{xspace}
\makeatletter
\DeclareRobustCommand\onedot{\futurelet\@let@token\@onedot}
\def\@onedot{\ifx\@let@token.\else.\null\fi\xspace}

\def\eg{\emph{e.g}\onedot} 
\def\ie{\emph{i.e}\onedot} 
 
\def\etc{\emph{etc}\onedot} \def\vs{\emph{vs}\onedot}
 
\def\etal{\emph{et al}\onedot}
\makeatother

\def\R{\mathbb{R}}

\usepackage[breaklinks=true,bookmarks=false,colorlinks=true]{hyperref}

\begin{document}

\title{End-to-end Temporal Action Detection with Transformer}

\author{Xiaolong Liu,~Qimeng Wang,~Yao Hu,~Xu Tang,~Shiwei Zhang,~Song Bai,~and~Xiang~Bai,~\IEEEmembership{Senior Member,~IEEE}
\thanks{This work was supported by National Key R\&D Program of China (No. 2018YFB1004600). \textit{(Corresponding author: Xiang Bai.)}}
\thanks{X. Liu (email: brucelio@outlook.com) and Q. Wang are with the School of Electronic Information and Communications, Huazhong University of Science and Technology. X. Bai (email: xbai@hust.edu.cn) is with the School of Artificial Intelligence and Automation, Huazhong University of Science and Technology. 
Y. Hu, X. Tang, and S. Zhang are with Alibaba Group. S. Bai is with ByteDance Inc.
Part of this work was done when X. Liu was an intern at Alibaba Group.}
\thanks{This paper has supplementary downloadable material available at http://ieeexplore.ieee.org., provided by the author.}
}

\markboth{Journal of \LaTeX\ Class Files,~Vol.~14, No.~8, August~2021}%
{Shell \MakeLowercase{\textit{et al.}}: A Sample Article Using IEEEtran.cls for IEEE Journals}

\IEEEpubid{0000--0000/00\$00.00~\copyright~2021 IEEE}

\maketitle

\begin{abstract}
Temporal action detection (TAD) aims to determine the semantic label and the temporal interval of every action instance in an untrimmed video. It is a fundamental and challenging task in video understanding. Previous methods tackle this task with complicated pipelines. They often need to train multiple networks and involve hand-designed operations, such as non-maximal suppression and anchor generation, which limit the flexibility and prevent end-to-end learning. In this paper, we propose an end-to-end Transformer-based method for TAD, termed TadTR. Given a  small set of learnable embeddings called action queries, TadTR adaptively extracts temporal context information from the video for each query and directly predicts action instances with the context. To adapt Transformer to TAD, we propose three improvements to enhance its locality awareness. The core is a temporal deformable attention module that selectively attends to a sparse set of key snippets in a video. A segment refinement mechanism and an actionness regression head are designed to refine the boundaries and confidence of the predicted instances, respectively. With such a simple pipeline, TadTR requires lower computation cost than previous detectors, while preserving remarkable performance. As a self-contained detector, it achieves state-of-the-art performance on THUMOS14 (56.7\% mAP) and HACS Segments  (32.09\% mAP). Combined with an extra action classifier, it obtains 36.75\% mAP on ActivityNet-1.3. Code is available at \url{https://github.com/xlliu7/TadTR}.
\end{abstract}

\begin{IEEEkeywords}
Transformer, Temporal Action Detection, Temporal Action Localization, Action Recognition.
\end{IEEEkeywords}

\section{Introduction}
\IEEEPARstart{V}{ideo} understanding has become more important than ever as the rapid growth of media prompts the generation, sharing, and consumption of videos. As a fundamental task in video understanding, temporal action detection (TAD) aims to predict the semantic label, the start time, and the end time of every action instance in an untrimmed and possibly long video. For its wide range of applications, including security surveillance, home care, video editing, video recommendation, and so on, temporal action detection has gained increasing attention from the community in recent years~\cite{shou2016temporal,xu2020g,Ma_2016_CVPR,richard2016temporal,caba2016fast}.

\begin{figure}
    \centering
    \includegraphics[width=\linewidth]{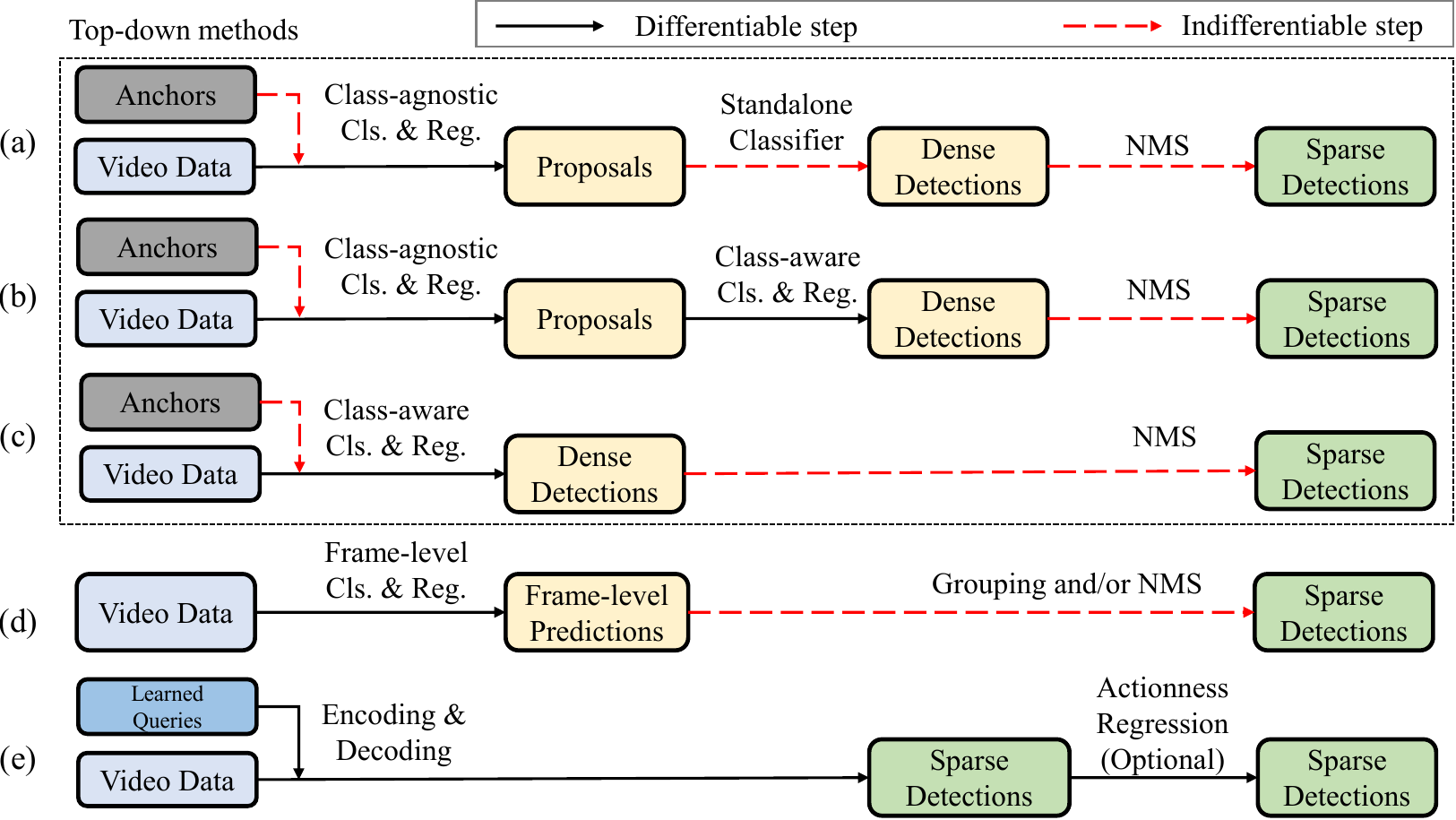}
    \caption{
    Comparison of different pipelines of temporal action detection. (a) Multi-stage pipeline in~\protect\cite{shou2016temporal,xu2020g},~\etc; (b) Two-stage pipeline in~\protect\cite{xu2017r,chao2018rethinking}; (c) Top-down one-stage pipeline in~\protect\cite{lin2017single}, (d) Bottom-up  pipeline in~\protect\cite{yuan2017temporal} (e) The set prediction pipeline in this work.}
    \label{fig:pipeline_compare}
\end{figure}

Previous methods for TAD can be roughly categorized into two groups. Top-down methods~\cite{shou2016temporal,zhao2017temporal,lin2019bmn} perform classification and regression on a large amount of  candidate segments. Bottom-up methods~\cite{yuan2017temporal,lea2017temporal} perform per-frame classification and group these predictions into segment-level predictions. While these methods achieve state-of-the-art performance on standard benchmarks, they  
have complex pipelines.

As shown in Fig.~\ref{fig:pipeline_compare}, these pipelines involve post-processing operations, such as non-maximum suppression (NMS) and grouping. These operations, together with anchor setting in many top-down methods, are hand-designed with prior knowledge about this task and not learnable, which restricts the flexibility.
Besides, most proposal-based methods~\cite{zeng2019graph,xu2020g,lin2019bmn}, requires a standalone classifier to classify action proposals. These issues block the gradient flow and prevent end-to-end learning.
Thus, it is necessary to develop a simple end-to-end method that directly predicts action instances in a single differentiable network\footnote{The single network means the detection network upon the video encoder. 
Training the video encoders along with the detection head for long videos often requires excessive computing resources. Therefore, most methods use offline features (\eg~I3D~\cite{carreira2017quo}) trained separately with large mini-batch size on large amounts of short (around 2 seconds) videos.} without hand-crafted components. 
\IEEEpubidadjcol

In this paper, we introduce an end-to-end temporal action detection framework to address the above issues. Inspired by the object detection Transformer (DETR)~\cite{carion2020end}, we directly map a set of learnable embeddings, called action queries, to action instances in parallel. 
As the queries do not directly indicate the initial locations of actions like anchors or proposals, we are unable to extract features for each query from specific locations like previous methods.
The detector is required to extract sufficient \textit{long-term context information} before knowing which interval an action falls in. Besides, the context should be \textit{adaptive} and relevant with each query, in order to differentiate between these queries. 
Traditional 1D convolutional neural networks cannot easily achieve them, due to a fixed receptive field and fixed weights. 
Recently, Transformers~\cite{vaswani2017attention} have shown great power in sequence modeling. It is able to reason the relations between sequence elements and adaptively capture long-term context with the self-attention module. A video is naturally a sequence of frames and there is abundant context in it~\cite{dai2017temporal,zeng2019graph,xu2020g}. Therefore Transformer is a desirable choice for the above goals.

Based on the above motivations, we propose a Temporal Action Detection TRansformer (TadTR) that predicts actions by extracting relevant context for each action query. It has an encoder-decoder structure. The encoder models inter-snippet relations to capture snippet-level context. The decoder models action-snippet relations to enhance each action query with snippet-level context, and inter-action relations to capture instance-level context from the other action queries. In this way, we can exploit richer context than previous methods that only exploit snippet-level context~\cite{xu2020g} or instance-level context~\cite{zeng2019graph}. Upon the decoder, two feed-forward networks (FFNs) predict the class and the segment for each action query. During training, an action matching module dynamically determines a one-to-one ground truth assignment according to the predictions. Owing to this, our detector avoids duplicate detections and NMS is unnecessary. It produces a very sparse set of action detections ($10 \sim 10^2$), orders of magnitude fewer than previous methods ($10^3 \sim 10^4$).

However, due to the intrinsic difference between space and time, a direct application of Transformer is not appropriate. We observe that different frames in a video of actions are highly similar, because of the temporal redundancy and the slow changes in backgrounds or actors. Besides, the boundaries of actions are less clear than those of objects~\cite{alwassel2018diagnosing}. Therefore, to precisely detect actions, a detector needs to be \textit{locality-aware}, which means being aware of the subtle local changes in the temporal domain. The dense attention module in primitive Transformer that attends to all elements in a sequence, is less sensitive to such local changes by design. 
To mitigate this issue, we draw inspiration from~\cite{zhu2021deformable} and propose a temporal deformable attention (TDA) module as the basic building block of Transformer.
It selectively attends to a sparse set of key elements around a reference location in the input sequence, where the sampling locations and attention weights are learned and dynamically adjusted in accordance with the inputs. In this way, it can adaptively extract context information while preserving locality awareness. 

Besides TDA, we make two additional improvements to enhance locality awareness. First, a segment refinement mechanism is employed to refine the boundaries of predicted actions. To be concrete, we iteratively re-attend to the video according to the previous predictions and refine the boundaries with the newly extracted context.
Second, we attach an actionness regression head to Transformer to predict a  reliable confidence score called actionness for detection ranking. 
It extracts the local features with RoIAlign~\cite{he2017mask} for each predicted action and estimates its IoU with the best-matched ground truth action. This is more reliable than simply using classification scores, as the classification branch may find a shortcut from context but ignore the complete local details.  
Despite being seemingly small changes, they significantly improve performance.

\begin{figure}[tb]
\centering
\includegraphics[width=\linewidth]{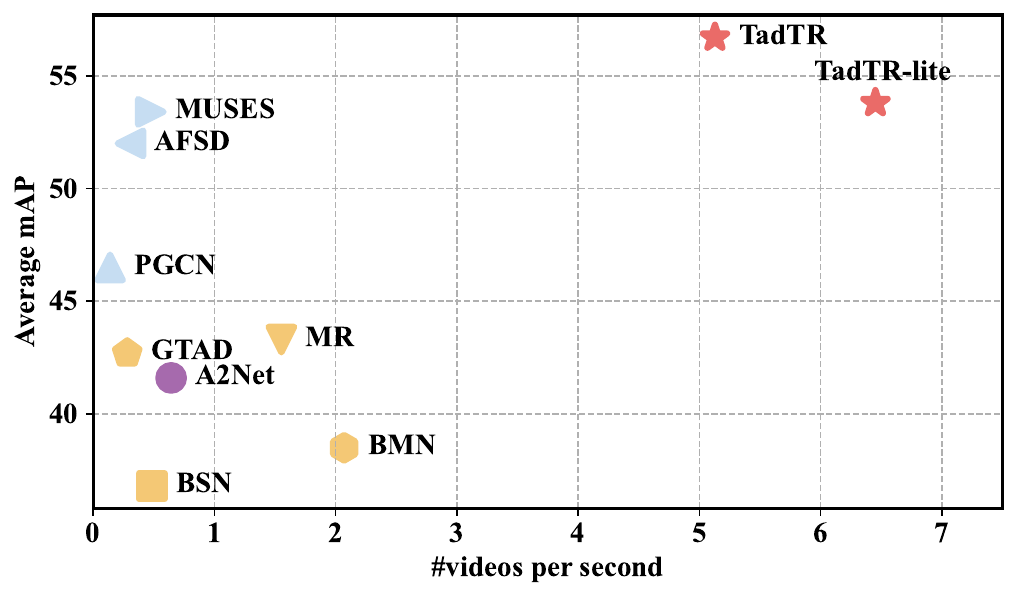}
\caption{
Comparison of recent temporal action detection methods on THUMOS14, in terms of both performance (average mAP) and speed. Our method achieves state-of-the-art performance while running significantly faster.
}
\label{fig:speed}
\end{figure}

We conduct comprehensive experiments on three datasets to evaluate TadTR. With a surprisingly simple pipeline, TadTR achieves remarkable performance with a low computation cost. Without any extra classifier, it achieves state-of-the-art performance on HACS Segments~\cite{zhao2019hacs} and THUMOS14~\cite{jiang2014thumos}. When combined an extra classifier, it reaches 36.75\% mAP on ActivityNet-1.3~\cite{caba2015activitynet}, outperforming strong competitors such as G-TAD~\cite{xu2020g} and BMN~\cite{lin2019bmn}.
In terms of run time, it takes only 155 ms per video on THUMOS14, which is much faster than  recent state-of-the-art methods, as shown in Fig.~\ref{fig:speed}.
We believe that the simplicity, the flexibility, and the strong performance of the new method will benefit and ease future research on temporal action detection.

The contributions of this work are as follows:
\begin{itemize}
    \item We introduce an end-to-end set prediction (SP) framework that simplifies the pipeline for temporal action detection (TAD). It can detect actions in a single differentiable network without hand-crafted components. 
    \item We propose a Transformer architecture that is enhanced with locality awareness to better adapt to the TAD task. The core is a temporal deformable attention (TDA) module that selectively attends to a sparse set of key snippets in a video. We show that TDA is crucial for the success of the SP framework for TAD.
    \item Different from previous works that ignore context or only exploit snippet-level or instance-level context, we model inter-snippet, inter-action, and action-snippet relations to capture both levels of context for more accurate temporal action detection.
    \item Our method achieves state-of-the-art performance of self-contained detectors on HACS Segments and THUMOS14, and competitive results on ActivityNet-1.3. Besides, it requires a lower computation cost than its competitors.
\end{itemize}

\section{Related Work}
\noindent\textbf{Temporal Action Detection.} Previous TAD methods can be roughly categorized into top-down methods and bottom-up methods according to the pipeline. Top-down methods can be further categorized into multi-stage, two-stage, and one-stage methods.
(a) \textbf{Multi-stage methods}~\cite{xu2020g,lin2019bmn,yuan2016temporal,heilbron2017scc,liu2020self,Liu_2021_CVPR} first generate candidate segments and train a binary classifier that associates each segment with a confidence score, resulting in proposals. Those proposals with high scores are fed to a multi-class classifier to classify the actions. 
The candidate segments are generated by dense uniform sampling~\cite{shou2016temporal,gao2017turn} or grouping local frames that may contain actions~\cite{lin2018bsn}.
Some methods~\cite{gao2018ctap,liu2019multi} combine multiple schemes for complementarity.
(b) \textbf{Two-stage methods}~\cite{xu2017r,chao2018rethinking,escorcia2016daps,buch2017sst} simplify the multi-stage pipeline by adopting a one-stage proposal generator, which directly predicts the scores and boundaries of pre-defined multi-scale anchors associated with each temporal location. 
These methods need to manually set multiple anchor scales, which restricts the flexibility. Note that multi-stage methods can also be seen as generalized two-stage methods. 
(c) \textbf{Top-down one-stage methods}~\cite{lin2017single,long2019gaussian} can be seen as the class-aware variant of the one-stage proposal generator. 
(d) \textbf{Bottom-up methods} perform frame-level action classification and merge the frame-level results to segment-level predictions. For example, \cite{yuan2017temporal} first predicts the action and boundary probabilities and then groups frames with maximal structured sum as actions. 
Recent anchor-free methods (\eg, AFSD~\cite{lin2021learning} and A2Net~\cite{yang2020revisiting}) also belong to this group. 
Besides these methods, a few works (\eg, CTAP~\cite{gao2018ctap} and PCG-TAL~\cite{su2021pcg}) combine different pipelines to enhance the performance. 
All the above methods require post-processing steps such as NMS or grouping, which prevent end-to-end learning.
An early work by Yeung~\etal~\cite{yeung2016end} also proposes a TAD method without hand-crafted components. Based on recurrent neural networks (RNN) and reinforcement learning (RL), it learns action detection by training an agent that iteratively picks an observation location and deciding whether to emit or refine a candidate action after observation. However, its reward function is not differetiable. Therefore it does not meet the criteria of end-to-end in this paper. 
All the above methods are fully-supervised. There are also some weakly-supervised methods that only utilize single-frame supervision~\cite{ma2020sf} or video-level supervision~\cite{nguyen2018weakly,paul2018w,liu2019completeness,shou2018autoloc,yu2019temporal,huang2021modeling,yang2021multi,zeng2019breaking,huang2020relational,islam2021hybrid,zhai2020two,huang2021foreground} during training. 

\vspace{1ex}\noindent\textbf{Transformers and Context in Video Understanding.}
Transformers have achieved great success in natural language processing~\cite{vaswani2017attention} and image understanding~\cite{tan2021planetr,chen2022transmix,liang2022transcrowd}. The core of Transformer is the self-attention mechanism that aggregates non-local cues through a weighted sum of features at attended locations. Compared with convolutions, self-attention can capture long-range context and dynamically adjust weights according to the input. Recently, many works have revealed the great potential of Transformers in video understanding tasks~\cite{sun2019videobert,bertasius2021space,wu2021seqformer}.
For example, VideoBERT~\cite{sun2019videobert} and ActBERT~\cite{zhu2020actbert} utilize Transformers to learn an joint representation for video and text. TimeSformer~\cite{bertasius2021space} decouples spatial and temporal self-attention for video classification. Zhou~\etal~\cite{zhou2018end} capture the temporal dependency with Transformer for video captioning.
Girdhar~\etal~\cite{girdhar2019video} apply Transformer to model the relationship between spatial proposals for spatio-temporal action detection.

In this paper, Transformer is used to capture temporal context information for temporal action detection.  Specifically, we employ attention modules to model the relations between video snippets, the relations actions and snippets, and the relations between actions. Several concurrent works also employ Transformer for context modeling in temporal action detection (AGT~\cite{nawhal2021activity}) and temporal action proposal generation (RTD-Net~\cite{tan2021relaxed} and TAPG~\cite{wang2021temporal}). 
However, these works either adopt a traditional TAD pipeline or have difficulty in training.
TAPG still relies on hand-crafted anchors and post-processing steps.  
RTD-Net requires a three-step training scheme to optimize different parts of the network separately and relies on extra action classifiers to classify the proposals. AGT suffers from slow training convergence (1000$\times$ more iterations than TadTR).
In addition, different from these works that exploit the vanilla attention module, TadTR introduces a more efficient temporal deformable attention module that adaptively attends to a sparse set of key snippets in a video. As a result, it enjoys lower computation costs and easier training. Therefore, TadTR is more practical. 

In the field of temporal action detection, some previous works also exploit context in other ways. For example, increasing the receptive field by a fixed ratio~\cite{chao2018rethinking,dai2017temporal}. However, this is not flexible enough and may introduce irrelevant information from unrelated frames.
Another line of works exploit context by modeling the relations between different snippets~\cite{xu2020g,bai2020boundary} or the relations between different proposals~\cite{zeng2019graph} with graph. The attention modules in this work are alternatives to them. Moreover, we model different kinds of relations and can capture richer context of different levels.

\vspace{1ex}\noindent \textbf{DETR and Deformable DETR. }
Temporal action detection methods~\cite{xu2017r,lin2017single,yang2020revisiting} often draw inspiration from object detection methods. This work is inspired by DETR~\cite{carion2020end} and Deformable DETR~\cite{zhu2021deformable}.
DETR proposes a Transformer-based Set Prediction (SP) framework to achieve end-to-end object detection without hand-crafted components.
Deformable DETR proposes multi-scale deformable attention to address the issues of slow convergence and limited feature resolution of DETR.
While extending them for direct TAD is intuitive, the effectiveness remains unclear. Our main contribution over DETR and Deformable DETR is that we adapt the SP framework and deformable attention for direct TAD and validate their effectiveness.
Although the high-level design of TadTR is similar to Deformable DETR, the implementation is different as TadTR aims to temporally localize actions in videos while Deformable DETR is designed for object detection in images. Besides, we reveal that deformable attention is crucial for the success of the SP framework for TAD and segment refinement is also important.
Furthermore, directly extending Deformable DETR to TAD does not achieve satisfactory performance as the confidence scores predicted by the decoder are not reliable. 
To relieve this issue, we add a simple yet effective actionness regression head to refine the confidence scores. To sum up, the adaptation and improvement over DETR and Deformable DETR make the SP framework practical for TAD and TadTR can serve as a strong baseline for SP-based TAD.


\section{TadTR}
\begin{figure*}
\centering
\includegraphics[width=0.9\linewidth]{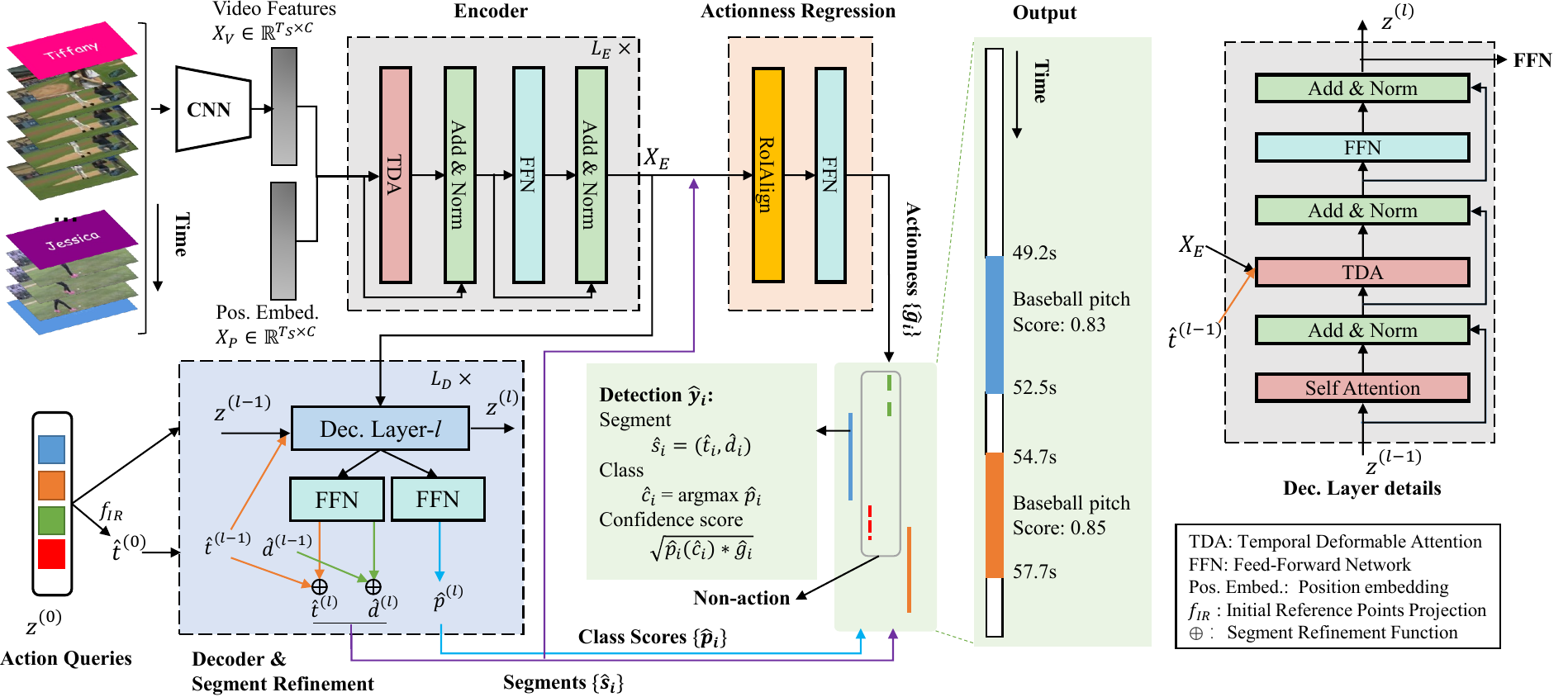}
\caption{The architecture of TadTR. It takes the video features extracted with a CNN and a set of learnable action queries as input and decodes a set of action predictions in parallel via a Transformer. The encoder captures the long-term context in the input feature sequence. The decoder extracts relevant context from the encoder for each action query and models the relations between action queries. Upon the decoder, we use feed-forward networks to predict the segments and classes of output actions. A segment refinement mechanism (the blue box) and an actionness regression head (the orange box) are utilized to refine the boundaries and the confidence scores of the predicted actions, respectively.}
\label{fig:arch}
\end{figure*}
TadTR is constructed on video features encoded with a pre-trained video classification network (\eg, I3D~\cite{carreira2017quo}).
Fig. \ref{fig:arch} shows the overall architecture of TadTR. TadTR takes as input the video features and a set of learnable action queries. Then it outputs a set of action predictions. Each action prediction is represented as a tuple of the temporal segment, the confidence score, and the semantic label. It consists of a Transformer encoder to model the interactions between video snippets, a Transformer decoder to predict action segments, and an extra actionness regression head to estimate the confidence score of the predicted segments. During training, an action matching module is used to determine a one-to-one ground truth assignment to the action predictions.

\subsection{Architecture}
\noindent\textbf{Encoder.}
Let $\boldsymbol{X}_V \in \R ^{T_S\times C}$ denotes the video feature sequence, where $T_S$ and $C$ are the length and dimension, respectively. Each frame in the feature sequence is a feature vector extracted from a certain snippet in the video. Here, a snippet means a sequence of a few (\eg, 8) consecutive frames. We use linear projection to make $C=256$.
The encoder models the relations between different snippets and outputs a feature sequence $\boldsymbol{X}_E \in \R^{T_S \times C}$ enhanced with temporal context. As depicted in Fig.~\ref{fig:arch}, it consists of $L_E$ Transformer encoder layers of the homogeneous architecture. Each encoder layer has two sub-layers,~\ie, a temporal deformable attention (TDA) module, and a feed-forward network (FFN). Layer normalization~\cite{ba2016layer} is used after each sub-layer and a residual connection is added between the input of each sub-layer and the output of the follow-up normalization layer. Except for TDA, all the other components are identical to the primitive Transformer~\cite{vaswani2017attention}.

TDA is an alternative to the dense attention module in~\cite{vaswani2017attention}. The high similarities between different frames and the vagueness of action boundaries require a detector to possess locality awareness. In other words, the detector should be more sensitive to local changes in the temporal domain. The dense attention module that attends to all locations in an input feature sequence, is less sensitive to such local changes. Besides, it suffers from high computation cost and slow convergence~\cite{zhu2021deformable}. 
To better fit the TAD task, we draw inspiration from~\cite{zhu2021deformable} and propose a temporal deformable attention (TDA) module that adaptively attends to a sparse set of temporal locations around a reference location in the input feature sequence.

Let $\boldsymbol{z}_q \in \R ^{C}$ be the feature of query $q$ and $t_q \in [0, 1]$ be the  normalized coordinate of the corresponding reference point. Given an input feature sequence $\boldsymbol{X}\in \R^{T_S\times C}$, the output $\boldsymbol{h}_m \in \R ^ {T_S \times (C/M)}$ of the $m$-th ($m \in \{1, 2, ..., M\}$) head of a TDA module is computed by an weighted sum of a set of key elements sampled from $\boldsymbol{X}$:  
\begin{equation} 
\label{eq:deform_attention}
    h_m = \sum_{k=1}^K a_{mqk} \boldsymbol{W}^{V}_m \boldsymbol{X}((t_q+\Delta t_{mqk})T_S),
\end{equation}
where $K$ is the number of sampling points, $a_{mqk} \in [0, 1]$ is the normalized attention weight, and $\Delta t_{mqk} \in [0, 1]$ is the sampling offset relative to $t_q$. $X((t_q+\Delta t_{mqk})T_S)$ is the linear interpolated feature at $(t_q+\Delta t_{mqk})T_S$ as it is fractional.  Following~\cite{zhu2021deformable}, the attention weight $a_{mqk}$ and the sampling offset $\Delta t_{mqk}$ are predicted from the query feature $\boldsymbol{z}_q$ by linear projection. We normalize the attention weight with \textit{softmax} to make $\sum_{k=1}^K a_{mqk}=1$.  $\boldsymbol{W}^{V}_m \in \R ^{C\times (C/M)}$ is a learnable weight. The output of TDA is computed by a linear combination of the outputs of different heads:
\begin{equation}
\label{eq:head}
    \text{TDA}(\boldsymbol{z}_q, t_q, X) = \boldsymbol{W}^{O} \text{Concat}(h_1, h_2, ..., h_m),
\end{equation}
where $\boldsymbol{W}^{O} \in \R ^{C\times C}$ is a learnable weight.

When computing the $\tau$-th frame in the output sequence, the query and the reference point are both the $\tau$-th frame in the input sequence. Therefore, we refer to TDA in the encoder as temporal deformable self-attention (TDSA). The query feature is the summation of the input feature of that frame and the position embedding at that location. The position embedding is used to differentiate between different locations in the input sequence.
In this paper, we use the sinusoidal position embedding following~\cite{vaswani2017attention}. 
\begin{equation}
    \boldsymbol{X}_P(\tau, \gamma) = \begin{cases} 
       \sin{\frac{\tau}{10000^{\gamma/C}}}& \gamma~\text{is even}\\
       \cos{\frac{\tau}{10000^{(\gamma-1)/C}}} & \gamma~\text{is odd}
    \end{cases}.
\end{equation}

The feed-forward network consists of two fully connected (FC) layers and a ReLU activation in between. It is the same across different positions and can be viewed as a stack of two 1D convolution layers with kernel size 1. The dimensions of the two FC layers are $C_F=2048$ and $C=256$, respectively. 

\vspace{1ex}\noindent\textbf{Decoder.}
The decoder takes as input the encoder features $\boldsymbol{X}_E $ and $N_q$ action queries with learnable embeddings $\hat{\boldsymbol{ z}}^{(0)}=\{\hat{\boldsymbol{ z}}_{i}^{(0)}\}_{i=1}^{N_q}$. It transforms these embeddings to $N_q$ action predictions $\hat{Y}=\{\hat{y}_i\}$. 
As illustrated in Fig.~\ref{fig:arch}, the decoder consists of $L_D$ sequential decoder layers. 
Each decoder layer has three major sub-layers: a self-attention module, a temporal deformable cross-attention (TDCA) module, and a feed-forward network. Similar to each encoder layer, we add a residual connection between each sub-layer and the following layer normalization function.  The output of the $l$-th decoder layer is denoted by $z^{(l)}$.
The self-attention module models the relation between action queries and updates their embeddings. The motivation here is that multiple actions in one video are often related. For example, a cricket shot action often appears after a cricket bowling action. To make an action prediction, each query extracts relevant context information from the video via the TDCA module. Given the encoder features $\boldsymbol{X}_E$ and the input embedding $\hat{\boldsymbol{z}}_{i} \in \R^{C}$, the output query embedding of TDCA is formulated as $TDA(\hat{\boldsymbol{z}}_{i}, \hat{t}_{i}, \boldsymbol{X}_E)$.  Here, $\hat{t}_{i}$ is the coordinate of the reference point in $\boldsymbol{X}_E$. By default, it is predicted by a projection function $f_{IR}$ from $\hat{\boldsymbol{ z}}_{i}^{(0)}$. $f_{IR}$ is implemented with a linear layer and a follow-up  \textit{sigmoid} function for normalization.
The reference point can be seen as the initial estimation of the center of the corresponding action segment. 
FFNs in the decoder layers have the same architecture as those in the encoder layers.

Different from TDSA in the encoder, the query embedding $\hat{\boldsymbol{z}}_{i}^{(0)}$ and the reference point are learnable and shared by all input videos. This allows the network to learn the global distribution of the action locations in the training dataset, which is more flexible than hand-crafted anchor setting or proposal sampling. An analysis is given in Sec.~\ref{subsec:exp:analysis}.

\vspace{1ex}\noindent\textbf{Prediction Heads.} Upon the output (the updated query embeddings) of each decoder layer, we apply FFNs to predict the classification probabilities $\hat{\boldsymbol{p}_i}$ and the temporal segment $\hat{s}_i=(\hat{t}_i, \hat{d}_i)$ of the action instance $\hat{y}_i$ corresponding to each query. Both $\hat{t}_i$ and $\hat{d}_i$ are normalized. To make the boundaries of the instances more accurate, a segment refinement mechanism is proposed. Besides, an additional actionness regression head is employed to refine the confidence score. They are detailed below.

\vspace{1ex}\noindent\textbf{Segment Refinement.} Transformer is able to capture long-range context information. However, the predicted action boundaries might be unsatisfactory for lack of locality. Inspired by~\cite{zhu2021deformable}, we introduce a refinement mechanism to enhance locality awareness and improve localization performance. 
It involves two strategies. The first is the incremental refinement of segments. Instead of predicting the segments independently at each decoder layer, we adjust the segments according to previously predicted segments layer by layer.
Formally, given each action segment $\hat{s}_i^{(l-1)}=(\hat{t}_i^{(l-1)}, \hat{d}_i^{(l-1)})$ predicted at the $(l-1)$-th decoder layer, the $l$-th decoder layer predicts the location offsets $(\Delta \hat{t}_i^{(l)}, \Delta \hat{d}_i^{(l)})$ relative to $\hat{s}_i^{(l-1)}$. The corresponding refined segment $\hat{s}_i^{(l)}=(\hat{t}_i^{(l)}, \hat{d}_i^{(l)})$ is then computed by:
\begin{align}
    \hat{t}_i^{(l)}&=\sigma(\Delta \hat{t}_i^{(l)}+ \sigma^{-1}(\hat{t}_i^{(l-1)})), l\in \{1, 2,...,L_D\}\\
    \hat{d}_i^{(l)}&=\sigma(\Delta \hat{d}_i^{(l)}+ \sigma^{-1}(\hat{d}_i^{(l-1)})), l\in \{2, 3,...,L_D\},
\end{align}
where $\sigma(\cdot)$ and $\sigma^{-1}(\cdot)$ are the sigmoid and the inverse sigmoid function, respectively.  Specially, $\hat{t}_i^{(0)}$, is the initial reference point $\hat{t}_i$ predicted by $f_{IR}$. The initial value of $\hat{d}_i^{(l)}$ is $\hat{d}_i^{(1)}$ predicted at the first decoder layer. 
The second is iterative reference point adjustment. We update the reference points of TDCA in each decoder layer instead of always using $\hat{t}_i^{(0)}$. Specifically, $\hat{t}_i^{(l-1)}$, the refined segment center at the $(l-1)$-th decoder layer, is used as the reference point of TDCA at the $l$-th decoder layer. 
In this way, TDCA can be adaptive to the input video and better aligned with the local features of the action instances. We validate the effectiveness of the two strategies in the experiments.

\begin{figure}
\centering
\includegraphics[width=\linewidth]{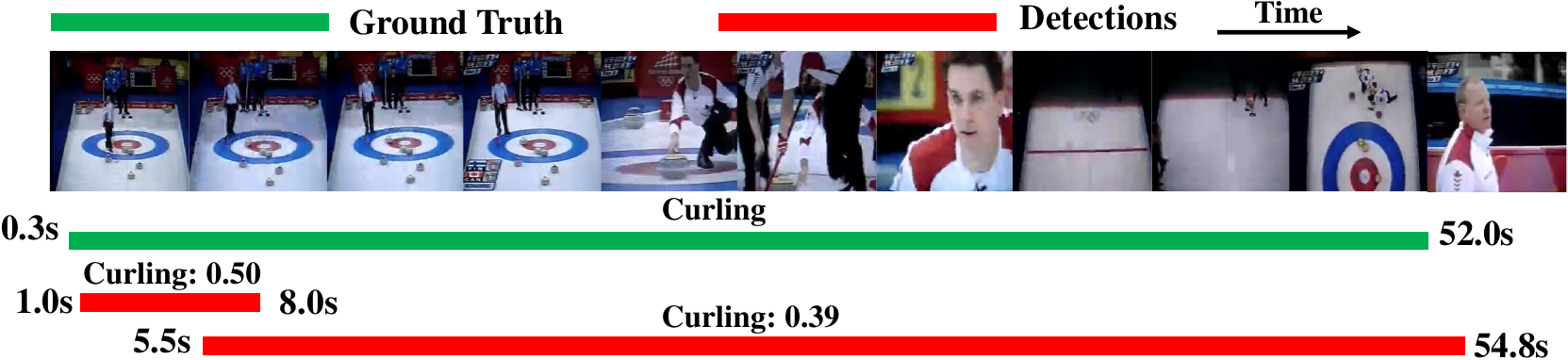}
\caption{The Transformer may generate unreliable confidence scores. Here, a prediction with a lower overlap with the curling action has a higher score than a more accurate prediction (0.50~\vs~0.39).}
\label{fig:scoring}
\end{figure}

\vspace{1ex}\noindent\textbf{Actionness Regression.} 
One challenge of temporal action detection is to generate reliable confidence scores for ranking. Typically, classification scores are used. However, the classification task focuses more on discriminative features and is less sensitive to the localization quality of an action.
As a result, the classification score of the detections may be unreliable for ranking. An example is shown in Fig.~\ref{fig:scoring}. 

To mitigate this issue, we employ an actionness regression head that extracts context aligned with the interval of a predicted segment and predicts an actionness score upon it. 
Given the encoder feature sequence $\boldsymbol{X}_E$ and a predicted segment $\hat{s}_i$ by the decoder, we first apply temporal RoIAlign~\cite{he2017mask} upon $\boldsymbol{X}_E$ to obtain the aligned features $\boldsymbol{X}_{s_i}\in\R^{T_R \times C}$ within the interval defined by $s_i$ from $\boldsymbol{X}_E$.
Here, $T_R$ is the number of bins for RoIAlign. 
To include a certain amount of context information around the boundaries, we slightly expand the segment by a factor of $\epsilon$ when applying RoIAlign. The expanded segment can be expressed as $(\hat{t}_i, \epsilon \hat{d}_i)$.
Then, a feed-forward network is used to predict the actionness score $\hat{g_i}$ from the aligned feature. $\hat{g_i}$ is supervised by the maximal IoU $g_i$ (intersection over union) between $s_i$ and all ground truth actions. In this way, the detector is enforced to be more sensitive to local features in order to differentiate between different segments.

\vspace{1ex}\noindent\textbf{Discussion.} The actionness regression head is somewhat similar to the second stage of the traditional two-stage method, as they can both refine the confidence scores. However, the set prediction pipeline adopted by TadTR is significantly different from traditional pipelines. It is hard to categorize TadTR and its variants into one-stage or two-stage methods, as it does not need anchor setting and post-processing steps like traditional one-stage or two-stage methods. 
Besides, the actionness regression head is very lightweight. It only needs to predict class-agnostic confidence scores. The computation cost (in FLOPs) of this head is 6.59\% of that of the full model.
Differently, the second stage of a two-stage method often contributes to a major amount of computation cost (\eg, 99.98\% in BMN~\cite{lin2019bmn}). We also note that TadTR can already make sparse and complete detections without actionness regression.

\subsection{Training and Inference}
\label{subsec:training}
\noindent\textbf{Action Matching.}
The action matching module determines the targets assigned to each detection during training. Inspired by DETR~\cite{carion2020end} in object detection, we frame it as a set-to-set bipartite matching problem to ensure a one-to-one ground truth assignment. 

Let $Y=\{y_j\}_{j=1}^{N_q}$ be a set of ground truth actions padded with $\varnothing$ (no action) and $\pi$ be the permutation that assigns each target $y_j$ to the corresponding detection $\hat{y}_{\pi(j)}$. Bipartie matching aims to find a permutation that minimizes the overall matching cost:
\begin{equation}
\label{eq:hungarian}
    \hat{\pi}=\arg \min \sum_{j=1}^{N_q} \mathcal{C}(y_j, \hat{y}_{\pi(j)}).
\end{equation}
The matching cost considers the classification probabilities and the distance between ground truth and predicated segments. Specifically, $\mathcal{C}(y_j, \hat{y}_{\pi(j)})$ is defined as
\begin{equation}
\label{eq:matching_cost}
    \mathbbm{1}_{c_j\neq\varnothing}[\mathcal{L}_{cls}(\boldsymbol{p}_{\pi(j)}, c_j)+\mathcal{L}_{seg}(s_j, \hat{s}_{\pi(j)})],
\end{equation}
where $c_j$ and $s_j$ are the class label and the temporal segment of $y_j$. $\mathcal{L}_{cls}(\boldsymbol{p}_{\pi(j)}, c_j)$ is the classification term. We use cross-entropy loss by default. $\mathcal{L}_{seg}(s_j, \hat{s}_{\pi(j)})$ is the distance between the predicted location and the ground truth location, defined as
\begin{equation}
     \lambda_{iou} \mathcal{L}_{iou}(s_j, \hat{s}_{\pi(j)}) + \lambda_{coord} \mathcal{L}_{L1}(s_j, \hat{s}_{\pi(j)}),
\end{equation}
where $\mathcal{L}_{L1}$ is the $L_1$ distance and $\mathcal{L}_{iou}$ is the IoU loss. IoU loss is defined as the the opposite number of the IoU.  $\lambda_{iou}$ and $\lambda_{coord}$ are hyper-parameters. The matching problem is solved with the Hungarian algorithm.

Through the set-based action matching, each ground truth will be assigned to only one prediction, thus avoiding duplicate predictions. This brings two merits. First, TadTR does not rely on the non-differentiable non-maximal suppression (NMS) for post-processing and enjoys end-to-end training. Second, we can make sparse predictions with limited queries (\eg~10) instead of dense predictions in many previous works (\eg tens of thousands for BMN~\cite{lin2019bmn} and G-TAD~\cite{xu2020g}), which saves the computation cost. 

In a way, the action matching module performs a learnable NMS. The matching cost takes the classification scores of the detections into account. In this way, those detections with lower scores are more likely to be assigned with a non-action target. As a result, their classification scores will be suppressed in the training process.

\vspace{1ex}\noindent\textbf{Loss Functions.} 
Once the ground truth assignment is determined, we optimize the network by minimizing the following multi-part loss functions:
\begin{equation}
\label{eq:loss_func}
\begin{aligned}
    \mathcal{L} = \sum_{j=1}^{N_q} [ \mathcal{L}_{cls}(\boldsymbol{p}_{\hat{\pi}(j)}, c_j) + \mathbbm{1}_{c_j\neq\varnothing}\mathcal{L}_{seg}(s_j, \hat{s}_{\hat{\pi}(j)}) \\
    + \lambda_{act}\mathcal{L}_{L1}(\hat{g}_{\hat{\pi}(j)},g_{\hat{\pi}(j)}) ],
\end{aligned}
\end{equation}
where the first two items optimize the detections from the decoder and the last one optimizes the outputs of actionness regression. $L_{cls}$ uses focal loss~\cite{lin2017focal}. $\hat{\pi}$ is the solution of Equation \ref{eq:hungarian}. $\lambda_{act}$ is a hyper-parameter.

\vspace{1ex}\noindent\textbf{Inference.} During inference, we ignore the action predictions from all but the last decoder layer. The confidence score for a detection $\hat{y}_i$ is computed by $\sqrt{\hat{\boldsymbol{p}}_i(\hat{c}_i) \cdot \hat{g}_i}$, where $\hat{c}_i$ is the predicted action label.


\section{Experiments}
\subsection{Experimental Setup}
\label{subsec:setup}
\noindent\textbf{Datasets and Evaluation Metrics.} We conduct experiments on THUMOS14~\cite{jiang2014thumos}, HACS Segments~\cite{zhao2019hacs}, and ActivityNet-1.3~\cite{caba2015activitynet}.
THUMOS14 is built on videos from 20 sports action classes. It contains 200 and 213 untrimmed videos for training and testing. There are 3007 and 3358 action instances on the two sets. The average length of actions is 5 seconds. ActivityNet-1.3 and HACS Segments share the same 200 classes of daily activities. Both datasets are split into three sets: training, validation, and testing. The numbers of videos in these sets are 10024, 4926, and 5044 respectively on ActivityNet-1.3, and 37613, 5981, and 5987 on HACS Segments. The average length of actions is 48 seconds on ActivityNet-1.3 and 33 seconds on HACS Segments.
On both datasets, the annotations on the testing set are reserved by the organizers. Therefore, we evaluate on the validation set.

Following conventions, the mean average precision (mAP) at different IoU thresholds is used for performance evaluation. On THUMOS14, the IoU thresholds for computing mAPs are $[0.3:0.7:0.1]$. On the other two datasets, we report mAPs at the thresholds $\{0.5, 0.75, 0.95\}$ and the average mAP at the thresholds $[0.5:0.95:0.05]$. For simplicity, we denote mAP at the IoU threshold $\alpha$ as mAP$_\alpha$ and the average mAP is referred to as mAP unless specially noted.

\vspace{1ex} \noindent \textbf{Video Feature Extraction.} 
Most TAD methods are based on offline extracted video features. For easier comparison with them, we also use video features as the input of our method. For experiments on HACS Segments, we directly use the official I3D features\footnote{\url{http://hacs.csail.mit.edu/hacs_segments_features.zip}}, which are extracted with I3D trained on Kinetics at 2FPS.
On the other datasets, we use the commonly used features in previous works. On THUMOS14, the two-stream I3D~\cite{carreira2017quo} networks pre-trained on Kinetics~\cite{carreira2017quo} are taken as the video encoder, and the features are extracted every 8 frames. On ActivityNet-1.3, we use the two-stream TSN~\cite{wang2016temporal} features extracted at 5FPS. Following previous works~\cite{lin2019bmn,xu2020g}, we resize the video features to a fixed length of 100 via linear interpolation on ActivityNet-1.3 and HACS Segments. Since the videos are long on THUMOS14, we follow~\cite{lin2019bmn} to crop each video feature sequence with windows of length 128 and stride 64 for training. In each window, we reserve the instances contained in it and clip the instances that partially overlap with it. We ignore the instances that have less than 1-second overlap with the window during training. This strategy is called length-based instance filtering. During inference, the stride is increased to 96 and the duplicate detections in the overlapped region are merged with NMS. This strategy is called cross-window fusion (CWF). We also report the performance of TadTR tested on non-overlapping windows (with a stride of 128). In this case, we simply take the union of detections from all windows of a video.

\vspace{1ex} \noindent \textbf{Implementation Details.}
$L_E$ and $L_D$ are set to 2 and 4, respectively.  The loss weights $\lambda_{iou}$, $\lambda_{coord}$ and $\lambda_{act}$ are set to 2, 5 and 5 respectively. 
The numbers of attention heads $M$ and sampling points $K$ are set to 8 and 4, respectively.
The parameters of the linear layers that predict attention weights are initialized to zero. We initialize the linear layers that predict sampling offsets to make $\{\Delta p_{mqk}\}_{m=1}^{8} =(k, 0, -k, 0, k, 0, -k, 0)$ at initialization. The expanding factor $\epsilon$ and the number of bins $T_R$ for RoIAlign in the actionness regression head are 1.5 and 16 respectively.

TadTR is trained using AdamW~\cite{loshchilov2017decoupled} optimizer. The initial learning rate is $2\times10^{-4}$ and scaled by a factor of 0.1 after training for a certain number of epochs. The learning rates of the linear projection layers for predicting attention weights and sampling offsets are multiplied by 0.1. We train the models for 30, 15, and 30 epochs and decrease the learning rates after 25, 12, and 25 epochs on THUMOS14, ActivityNet-1.3, and HACS Segments respectively. 
The batch size is set to 16.

\begin{table*}[tb]
\caption{Comparison with state-of-the-art methods on THUMOS14. Run time is the average inference time per video, including post-processing operations, such as NMS. SN: single-network. E2E: end-to-end. TS: two-stream. $^\sharp$For proposal generation methods, the computation cost of the extra classifiers is not included (marked with >). $\dagger$ Results copied from~\cite{yang2020revisiting}. $\ddagger$ Our implementation. * With focal loss and IBIF. $^\S$ With IBIF.}
\label{tab:thumos14_comparison}
\centering
\begin{tabular}{l|c|p{0.5cm}<\centering|c<{\centering}|*{6}{p{0.75cm}<{\centering}}|cc}
\toprule
Method&Feature&SN&E2E&mAP$_{0.3}$ &mAP$_{0.4 }$&mAP$_{0.5}$&mAP$_{0.6}$ & mAP$_{0.7}$ & mAP&{\scriptsize Time/ms} &GFLOPs  \\
\midrule
Yeung~\etal~\cite{yeung2016end}&VGG16&\cmark&-& 36.0&26.4& 17.1 &-&-&-&-&-\\
Yuan~\etal~\cite{yuan2017temporal}&TS &-&- &36.5 &27.8 &17.8& - &-&-&-&-\\
SSAD~\cite{lin2017single}&TS&\cmark&-&43.0&35.0&24.6&-&-&-&-&-\\
R-C3D~\cite{xu2017r}&C3D&\cmark&-&44.8&35.6&28.9&-&-&-&-&-\\
SSN~\cite{zhao2017cuhk}&TS&-&-&51.9&41.0&29.8&-&-&-&-&-\\
TAL-Net~\cite{chao2018rethinking}&I3D&\cmark&-&53.2&48.5&42.8&33.8&20.8&39.8&-&-\\
BSN~\cite{lin2018bsn}&TS&-&-&53.5&45.0&36.9&28.4&20.0&36.8&>2065&>3.4\\
MGG~\cite{liu2019multi}&TS&-&-& 53.9&46.8&37.4&29.5&21.3&37.8&-&-\\
BMN~\cite{lin2019bmn}&TS&-&-&56.0&47.4&38.8&29.7&20.5&38.5&>483&>171.0\\
BC-GNN~\cite{bai2020boundary}&TS&-&-&57.1&49.1&	40.4&	31.2&	23.1&	40.2&-&-\\
G-TAD~\cite{xu2020g} &TS&-&-&54.5&47.6&40.2&30.8&23.4&39.3&>4440&>639.8\\
BMN$^\dagger$ ~\cite{lin2019bmn}&I3D&-&-&56.4& 47.9& 39.2& 30.2& 21.2&39.0&-&-\\
G-TAD$^\ddagger$ ~\cite{xu2020g}&I3D&-&-&58.7&52.7&44.9&33.6&23.8&42.7&>3552&>368.9\\
MR~\cite{zhao2020bottom} &I3D&-&-&53.9&50.7&45.4&38.0&28.5&43.3&>644&>36.8\\
A2Net~\cite{yang2020revisiting}&I3D&\cmark&-&58.6&54.1&45.5&32.5&17.2&41.6&1554&30.4\\
P-GCN~\cite{zeng2019graph}&I3D&-& -&63.6&57.8&49.1&-&-&-&7298&4.4\\
P-GCN$^\ddagger$~\cite{zeng2019graph}&I3D&-&-&64.9&	59.0&	49.4&	36.7&22.6&46.5&7298&4.4\\
G-TAD~\cite{xu2020g}+P-GCN~\cite{zeng2019graph}&I3D&-&-&66.4& 60.4& 51.6 &37.6& 22.9&47.8&-&-\\
AGT~\cite{nawhal2021activity}&I3D&\cmark&\cmark& 65.0& 58.1& 50.2&-&-&-&-&-\\
PCG-TAL~\cite{su2021pcg}&I3D &-&-&64.2&57.3&48.3&-&-&-&-&-\\
RTD-Net~\cite{tan2021relaxed} &I3D&-&-&68.3&62.3&51.9&38.8&23.7&	49.0&>211&>32.1 \\
AFSD~\cite{lin2021learning}&I3D&-&-&67.3&	62.4&	55.5&	43.7&	31.1&	52.0&3245 &84.1\\
MUSES~\cite{Liu_2021_CVPR}&I3D&-&-&68.9 &64.0& 56.9& 46.3&31.0& 53.4&2101& 34.1\\

\hline
TadTR* (Ours) &I3D&\cmark&\cmark&\textbf{74.8}& \textbf{69.1}& \textbf{60.1}& \textbf{46.6}& \textbf{32.8}& \textbf{56.7} &195&1.07 \\ 
TadTR-lite* (Ours)&I3D&\cmark&\cmark&71.3& 65.9&57.0& 44.6& 30.4& 53.8&\textbf{155}&\textbf{0.85} \\
TadTR$^\S$ (Ours)&I3D&\cmark&\cmark&70.3& 64.3& 55.7& 44.0& 30.0 &52.9& 195& 1.07\\
TadTR (Ours) &I3D&\cmark&\cmark& 67.1&61.1&52.0&39.9&26.2&49.3&195&1.07\\
\bottomrule
\end{tabular}
\end{table*}

In the experiments, we also explore an improved training setting. Following RetinaNet~\cite{lin2017focal} and AFSD~\cite{lin2021learning}, we use focal loss~\cite{lin2017focal} for the classification term in the matching cost (Eq.~\ref{eq:matching_cost}). We find that this modification speeds up convergence. Therefore the total numbers of training epochs are reduced to 16, 12, and 20 on the three datasets, respectively. The learning rates are decreased after 14, 9, and 18 epochs, respectively.  Besides, we use the integrity-based instance filtering (IBIF) strategy in G-TAD~\cite{xu2020g} and AFSD~\cite{lin2021learning} to replace the default length-based instance filtering strategy on THUMOS14. To be specific, in each window, we only keep those ground truth instances whose integrity exceeds 0.75. Here, the integrity of an instance $s_g$ in a window $s_w$ is defined as $|s_g \cap s_w|/|s_g|$, where $|\cdot|$ means the length. It is similar to IoU but has a different denominator. Those windows without such instances are ignored during training.

The experiments are conducted on a workstation with a single Tesla P100 GPU card, and Intel(R) Xeon(R) CPU E5-2682 v4 @ 2.50GHz. It takes around 10 minutes, 36 minutes, and 150 minutes to finish training on THUMOS14, ActivityNet-1.3, and HACS Segments, respectively.

\subsection{Main Results}

\noindent\textbf{THUMOS14.} Table~\ref{tab:thumos14_comparison} demonstrates the temporal action detection performance and run time comparison on the testing set of THUMOS14. We measure the run time of these methods with publicly available implementations under the same environment (a single P100 GPU). We run methods on the full testing set with batch size set to 1 and report the average time and FLOPs per video. The average length of videos on THUMOS14 is 217 seconds. BMN~\cite{lin2019bmn} and G-TAD~\cite{xu2020g} use two-stream TSN~\cite{wang2016temporal} features originally. For a fair comparison, we also report their performance with I3D features. For AFSD~\cite{lin2021learning}, we have excluded the computation cost of the feature extractor. For TadTR, we report the performance with different training and inference settings. 
The entry with $\S$ is with integrity-based instance filtering (IBIF). The entries with $*$ are with IBIF and focal loss. TadTR-lite is the variant that does not use cross-window fusion (CWF) during inference.
We observe that:

\noindent 1) TadTR* achieves the best performance among all the compared methods in terms of mAP at all IoU thresholds. Even the variant without CWF can achieve state-of-the-art performance. TadTR* is slightly better than TadTR$^\S$ owing to focal loss.

\noindent 2) TadTR* outperforms the second-best method MUSES~\cite{Liu_2021_CVPR} by 3.3\% in terms of average mAP, which demonstrates the advantage of our method.
Compared with the competitive single-network method A2Net~\cite{yang2020revisiting}, TadTR achieves 15.1\% higher average mAP. 

\noindent 3) Compared with the concurrent Transformer-based methods AGT~\cite{nawhal2021activity} and RTD-Net~\cite{tan2021relaxed}, TadTR* achieves better performance. It surpasses AGT by 9.7\% in terms of mAP$_{0.5}$. It also outperforms RTD-Net by 7.7\% (56.7\%~\vs~49.0\%) in terms of average mAP. Besides the advantage in accuracy, TadTR is easier to train. Differently, AGT requires 1000$\times$ more training iterations (3000k~\vs~3k for TadTR) due to slow convergence of dense attention. RTD-Net requires a three-step training scheme to optimize different parts of the network.

\noindent 4) Our results are achieved at a low computation cost. 
TadTR is around 11$\times$ faster than the second-best method MUSES, 
and 8$\times$ faster than the competitive single-network detector A2Net. It also requires much fewer FLOPs. 
The efficiency of our method is owing to the simple framework and the sparsity of predictions.

The above results indicate that our method is both accurate and efficient. We also note that many other methods are composed of multiple independently trained networks. Those proposal generation methods (BSN, MGG, BMN, G-TAD, MR, BC-GNN, and RTD-Net) are not self-contained, as they rely on an extra classifier (such as P-GCN) to accomplish the TAD task. Differently, TadTR can achieve action detection with only a \textit{single} unified network.

We note that the computation cost of TadTR is  not comparable with those methods that directly take video frames as input, such as R-C3D. Most previous works use video features as input and focus on the design of detection networks.

\vspace{1ex}\noindent\textbf{HACS Segments.} 
We report the performance of TadTR, SSN~\cite{zhao2017temporal}, and the state-of-the-art method G-TAD~\cite{xu2020g} in Table~\ref{tab:hacs_sota}. 
Our method achieves an average mAP of 30.83\%, which outperforms SSN (+11.86\% mAP) and G-TAD (+3.35\% mAP). 
Besides, our method requires 455$\times$ fewer GFLOPs than G-TAD. As for run time, the network inference and post-processing step of G-TAD take 33 ms and 908 ms per video, respectively. The total run time is 941 ms, 49.5$\times$ that of TadTR (19 ms). 
We also try the improved training setting, which results in 32.09\% mAP.
The results again illustrate the superiority of TadTR.

\begin{table}[tb]
\caption{
Comparison of different methods on the validation set of HACS Segments. The results of SSN are from \protect\cite{zhao2019hacs}. * With focal loss.}
\label{tab:hacs_sota}
\scriptsize
\centering
\begin{tabular}{l|*{4}{p{0.7cm}<{\centering}}|cc}
\toprule
Method& \scriptsize{mAP$_{0.5}$}&\scriptsize{mAP$_{0.75}$}& \scriptsize{mAP$_{0.95}$}&\scriptsize{mAP}&\scriptsize{Time/ms}&GFLOPs \\
\midrule
SSN~\cite{zhao2017cuhk} & 28.82& 18.80& 5.32& 18.97&-&-\\
G-TAD~\cite{xu2020g} & 41.08&27.59&8.34&27.48&941&45.7 \\
TadTR  & 45.16& 30.70& \textbf{11.78}& 30.83 &\textbf{19} &\textbf{0.1}\\
TadTR* & \textbf{47.14}& \textbf{32.11}& 10.94& \textbf{32.09}&\textbf{19} &\textbf{0.1}\\
\bottomrule
\end{tabular}
    
\end{table}

\begin{table*}[tb]
\caption{Comparison of different methods on ActivityNet-1.3. 
Methods in the second group are combined with an ensemble of action classifiers~\protect\cite{zhao2017cuhk}. The computation costs (in FLOPs) of the action classifiers are not included. The results of BMN and G-TAD with TSP~\cite{alwassel_2021_tsp} features are from~\cite{alwassel_2021_tsp}. TS: two-stream. SN: single-network. * With focal loss.
}
\label{tab:anet_sota}
\centering
\begin{tabular}{l|c|p{0.8cm}<\centering|p{0.8cm}<\centering|*{4}{p{1.1cm}<{\centering}}|c}
\toprule
Method & Feature&SN & E2E&mAP$_{0.5}$&mAP$_{0.75}$ &mAP$_{0.95}$ &mAP &GFLOPs\\
\midrule
\multicolumn{9}{l}{\textit{\textbf{Self-contained methods}}} \\ \hline
R-C3D~\cite{xu2017r}&C3D &\cmark& -&26.80 &- &- &- &- \\
SSN~\cite{zhao2017temporal}&TS& -&- &39.12& 23.48 &5.49& 23.98&-\\
TAL-Net~\cite{chao2018rethinking}&I3D& \cmark&-& 38.23& 18.30& 1.30&20.22&-\\
P-GCN~\cite{zeng2019graph} &I3D&- &- &42.90 &28.14& 2.47& 26.99&5.0 \\
PCG-TAL~\cite{su2021pcg}&I3D&-&-&42.14&28.34&6.12&27.34&-\\
TadTR (Ours) & TS&\cmark&\cmark& 41.40 & 28.85& 7.86& 28.21& \textbf{0.038}\\
TadTR* (Ours)& TS& \cmark&\cmark& \textbf{43.67}& \textbf{30.58}& \textbf{8.32}& \textbf{29.90}&\textbf{0.038}\\
   
\midrule
\multicolumn{9}{l}{\textbf{\textit{Combined with an ensemble of action classifiers}~\cite{zhao2017cuhk}}} \\ \hline
CDC~\cite{shou2017cdc}&C3D & -& -& 43.83& 25.88& 0.21& 22.77&-\\
BMN~\cite{lin2019bmn}&TS&-&-&50.07& 34.78& 8.29& 33.85&45.6\\
G-TAD~\cite{xu2020g}&TS&-&-& 50.36& 34.60&9.02 &34.09&45.7\\
P-GCN~\cite{zeng2019graph}&I3D & -&- &48.26& 33.16& 3.27& 31.11&5.0 \\
MR~\cite{zhao2020bottom}&I3D&-&-&43.47 &33.91& 9.21& 30.12&-\\
A2Net~\cite{yang2020revisiting}&I3D&-&-&43.55&28.69&3.70&27.75&1.2\\
PCG-TAL~\cite{su2021pcg}&I3D&-&-&50.24&35.21&7.84&34.01&-\\
RTD-Net~\cite{tan2021relaxed}&I3D& -&-&47.21&	30.68&	8.61&	30.83&3.1\\
AFSD~\cite{lin2021learning}&I3D&-&-& 52.38&	35.27&6.47&	34.39&3.3\\
TadTR* (Ours)&TS&-&-&51.29&34.99& 9.49 &34.64 & \textbf{0.038}\\
TadTR+BMN (Ours)&TS &-&-& 50.51&35.35&8.18&34.55&45.6\\
TadTR* (Ours)& I3D& -&-&\textbf{52.83}& \textbf{37.05}& \textbf{10.83} &\textbf{36.11}&\textbf{0.038}\\
\hline
BMN~\cite{lin2019bmn} &TSP & -&-&51.23	&36.78	&9.50	&35.67&45.6\\
G-TAD~\cite{xu2020g} &TSP & -&-&51.26&	37.12&	9.29&	35.81&45.7\\
TadTR*  (Ours) &TSP& - & - & \textbf{53.62}& \textbf{37.52}& \textbf{10.56}& \textbf{36.75}&\textbf{0.038}\\
\bottomrule
\end{tabular}
\end{table*}

\begin{table*}[tb]
\caption{
Comparison of different variants of TadTR.}
\label{tab:hacs_ablation}
\centering
 \begin{tabular}{l|*{4}{p{0.8cm}<{\centering}}c|c|c}
\toprule
\multirow{2}{*}{Method}&\multicolumn{5}{c|}{HACS Segments} & \scriptsize{THUMOS14} & \scriptsize{ActivityNet} \\ \cline{2-6}
&mAP$_{0.5}$&mAP$_{0.75}$ &mAP$_{0.95}$ &mAP&MFLOPs &mAP &mAP\\
\midrule
TadTR  & \textbf{45.16}& \textbf{30.70}& \textbf{11.78}& \textbf{30.83}  &100.5& \textbf{47.92}&\textbf{28.21}\\
TadTR w/o encoder & 39.65 &26.99& 9.08& 26.94& 95.3 &40.99&27.34\\
TadTR    w/o instance-level context & 43.11 &29.97 &10.43& 29.70&\textbf{66.8}&45.26&26.23\\
TadTR with dense attention & 22.76& 12.52& 4.19 &13.58&564.5&24.15&23.79\\
TadTR    w/o actionness regression  & 42.10& 28.44& 10.23& 28.51&99.4 &45.09&26.13\\
TadTR    w/o segment refinement & 39.89& 28.03& 9.54& 27.65&99.8&44.07&27.40 \\
\bottomrule
\end{tabular}
\end{table*}

\vspace{1ex}\noindent\textbf{ActivityNet-1.3.} 
Table~\ref{tab:anet_sota} compares the performance of different methods on the validation set of ActivityNet-1.3. 
Some methods (\eg, G-TAD~\cite{xu2020g}) only implement action proposal generation and cannot produce action detections without external action classifiers. 
We divide the methods into two groups according to whether external action classifiers are used. Being simple and end-to-end trainable, TadTR achieves an average mAP of 28.21\%, which is stronger than all the other methods. With the improved training setting, TadTR achieves 1.69\% higher mAP.  This variant is 2.56\% better than the second-best method PCG-TAL~\cite{su2021pcg} in terms of mAP. Compared with the second-best single-network detector TAL-Net, we improve the performance by 9.68\%.  

For comparison with previous methods~\cite{xu2020g,lin2019bmn,yang2020revisiting} that are combined with an ensemble of classifiers~\cite{zhao2017cuhk}, we also try such a combination. To be concrete, we pass the detections by TadTR to the classifiers and fuse the classification scores of TadTR and the classifiers by multiplication. 
When fused with~\cite{zhao2017temporal}, TadTR enjoys a significant performance boost, achieving an average mAP of 34.64\%. 
It is better than the other compared methods in terms of average mAP, although some methods use the stronger I3D features. When using the I3D features, the average mAP is further boosted to 36.11\%, outperforming the second-best method AFSD by 1.72\%. Besides, the performance is achieved at a low computation cost, as indicated by the smaller FLOPs.

TadTR can also be combined with BMN. This is implemented by connecting the encoder of TadTR to the detection head of BMN. Owing to the adaptive context captured by the Transformer encoder, TadTR+BMN achieves an improvement of 0.7\% over BMN, reaching 34.55\% mAP.  It also outperforms the recent method G-TAD. This indicates the advantage of Transformer in temporal action detection.
Note that 99.98\% of the computation cost is on the proposal classification branch of BMN. Therefore the total computation cost of TadTR+BMN is close to that of BMN.

With the stronger TSP~\cite{alwassel_2021_tsp} features, the performance of TadTR reaches 36.75\% mAP. It outperforms G-TAD by 0.94\% and BMN by 1.08\%. This again demonstrates the superiority of TadTR.

\subsection{Ablation Study}
In this subsection, we validate the effectiveness of different components of TadTR and evaluate the effects of various hyper-parameters. Unless specially noted, all reported results are with the default training and setting. TadTR-lite is used for THUMOS14.

\vspace{1ex}\noindent\textbf{The importance of context information.} The key of Transformer is the self-attention mechanism that incorporates the context in a video sequence. In TadTR, we leverage two kinds of context, snippet-level context from related snippets and instance-level context from related action queries, which are captured by Transformer encoder and the self-attention module in Transformer decoder respectively. 
By removing Transformer encoder, we get a variant ``TadTR w/o encoder''.  By removing the self-attention module in the decoder, we get a variant ``TadTR w/o instance-level context''.
We report the performance of the two variants in Table~\ref{tab:hacs_ablation}. It is observed that removing the encoder leads to a 3.89\% drop on HACS Segments, 6.93\% drop on THUMOS14, and 0.87\% drop on ActivityNet in terms of average mAP. It indicates that the Transformer encoder is crucial for our model, as the decoder requires long-range and adaptive context to reason the relations between the actions and the video. 
Removing instance-level context, the average mAP drops by 1.13\% on HACS Segments, 2.66\% on THUMOS14, and 1.98\% on ActivityNet. We conclude that the context information from other action instances is also helpful for action detection.

\begin{table}[tb]
\caption{Comparison of the variants of TadTR with Transformer encoder and 1D CNN encoder on HACS Segments. 
}
\label{tab:hacs_encoder}
\centering
\begin{tabular}{p{2.2cm}|*{2}{p{1.6cm}<{\centering}}|p{1.6cm}<{\centering}}
\toprule
\multirow{2}{*}{Encoder}&\multicolumn{2}{c|}{Average mAP}&\multirow{2}{*}{MFLOPs} \\ \cline{2-3}
& w/o NMS & w/ NMS & \\
\midrule
Transformer & 30.83 & 30.53& 100.5\\
1D CNN &27.95 &29.12&134.8\\
\bottomrule
\end{tabular}
\end{table}

\begin{figure}[tb]
    \centering
    \includegraphics[width=0.6\linewidth]{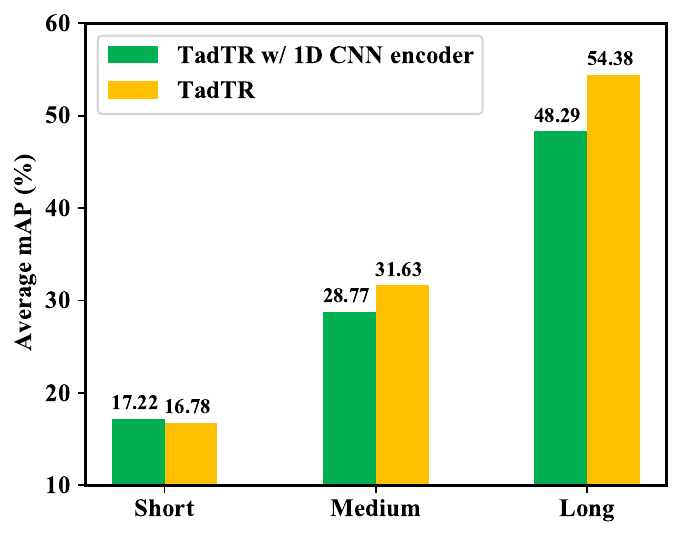}
    \caption{Comparison of the performance of TadTR with 1D CNN encoder and TadTR (with Transformer encoder) for actions with different durations on HACS Segments, measured by average mAP.}
    \label{fig:ap_by_length}
\end{figure}

\vspace{1ex}\noindent\textbf{Transformer encoder~\textit{v.s.}~CNN encoder.}
We try replacing the Transformer encoder with a 1D CNN encoder, which is common for temporal modeling in previous TAD methods. The 1D CNN encoder is composed of two 1D convolutional layers with 256 filters of kernel size 3 and ReLU activation. As can be observed in Table~\ref{tab:hacs_encoder}, using 1D CNN encoder leads to 2.88\% average mAP drop when NMS is not applied (the default option), and 1.41\% average mAP drop when NMS is applied. 
Interestingly, the performance of this variant with NMS is improved by 1.17\% over that without NMS. It indicates that there are many duplicate detections. One possible reason is that CNN features are locally correlated. Therefore, it is hard to differentiate between close predictions as they have similar features. 

To further dissect the performance gap between TadTR with 1D CNN encoder (equipped with NMS) and TadTR, we divide the ground truth instances into 3 groups according to the normalized duration: short ($0\sim0.1$), medium ($0.1\sim0.2$) and long ($0.2\sim1$) and report the average mAP for each group in Fig.~\ref{fig:ap_by_length}. 
Here we use the normalized duration because we resize the video features into a fixed length. For reference, the average duration per video is 148 seconds. As can be observed, TadTR with Transformer encoder achieves better performance for medium-length and long actions. 1D CNN encoder is slightly better for short actions. The result is reasonable, as 1D CNN is 
good at modeling short-term dependency but poor at modeling long-term dependency.

We also explore deeper CNNs and larger convolution kernels, but no improvement is observed.
In terms of computation cost, this variant has much higher FLOPs than TadTR with Transformer encoder. 
The results show that 1D CNN is inferior to Transformer encoder.

\vspace{1ex}\noindent\textbf{Dense attention \textit{v.s.} temporal deformable attention.} We try replacing all temporal deformable attention modules in TadTR with vanilla dense attention modules. This variant is called ``TadTR with dense attention''. As depicted in Table~\ref{tab:hacs_ablation}, the performance of this variant is far behind TadTR, especially on THUMOS14 (-23.77\% mAP), even if the model is trained for 180 epochs. It means temporal deformable attention is crucial for the success of TadTR.
The main reason is that the dense attention lacks locality awareness. 
As different frames are usually similar in background, a dense attention module tends to over-smooth input sequence at initialization (see an example in the supplementary material). As a result, it is hard to localize temporal segments with different semantics. 
Besides, the variant with dense attention has 5.7$\times$ higher computation cost than that of TadTR.
Therefore, temporal deformable attention is a better choice.

\begin{figure}[tb]
\centering
\includegraphics[width=\linewidth]{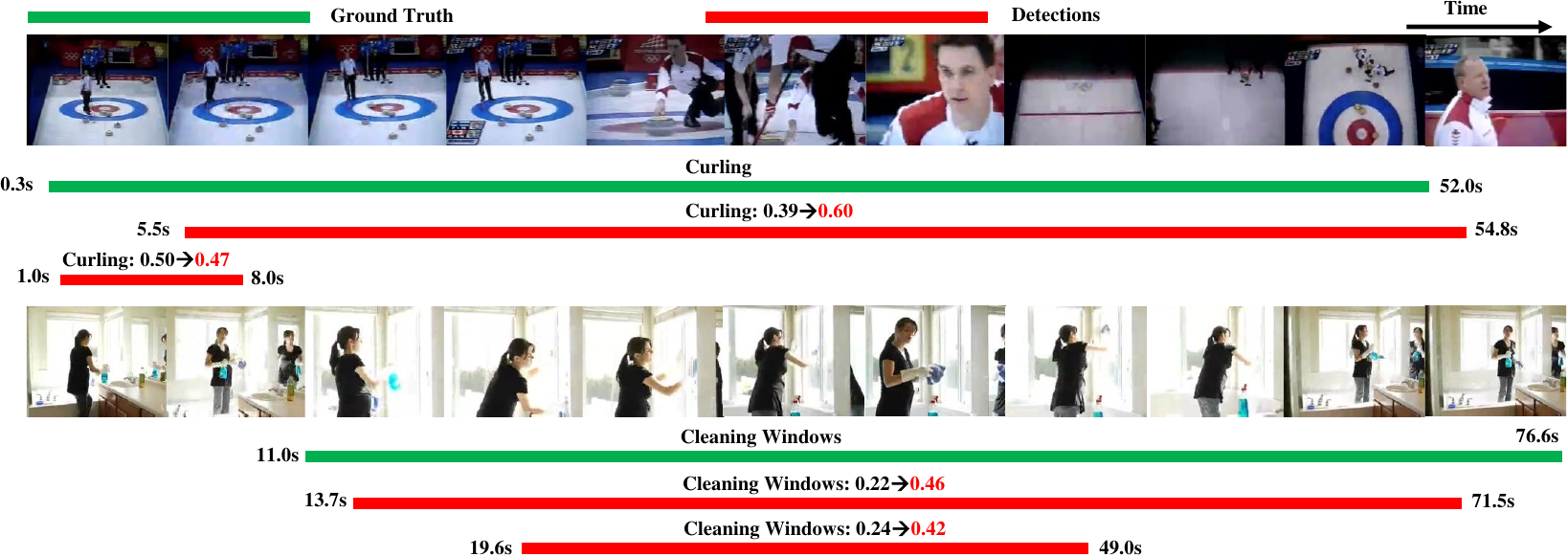}
\caption{Actionness regression improves the ranking of detections. In each of the above cases, the initial scores are unreliable. The more accurate detection obtains a lower score than the less accurate one before rescoring. With the new scores (on the right of the arrows) applied, the ranking order turns satisfactory. Best viewed in color.
}
\label{fig:qualitative}
\end{figure}

\vspace{1ex}\noindent\textbf{The effects of actionness regression and segment refinement.}
We study the effects of the two components by removing them individually, resulting in 2 different model variants. Their results are presented in Table~\ref{tab:hacs_ablation}. 
Comparing TadTR w/o actionness regression and TadTR, we observe that  actionness regression leads to improvements of all metrics. 
Specifically, the improvements are 3.06\%,  2.26\%, 1.55\%, and 2.32\% in terms of mAP at IoU 0.5, 0.75, 0.95 and the average mAP on HACS Segments. On THUMOS14 and ActivityNet, the average mAP improves by 2.83\% and 2.08\%. Several qualitative examples are presented in Fig.~\ref{fig:qualitative} to show how actionness regression helps. It can produce more reliable confidence scores for the action predictions.

Comparing TadTR w/o segment refinement and TadTR, we find this component helpful for improving the localization accuracy. On HACS Segments, it improves the mAP at the strict threshold of 0.95 by a large margin of 2.24\%. The mAPs at other thresholds are also consistently improved. 
On THUMOS14 and ActivityNet, the average mAP improves by 3.85\% and 0.81\%.

\begin{table}[tb]
\centering
\caption{Ablation study of segment refinement (SR) on ActivityNet. The results are with focal loss.}
\label{tab:segment_refine_ablation}
\begin{tabular}{l|cccc}
\toprule
Variants & 0.5&0.75&0.95&mAP \\
\midrule
Standard SR & \textbf{43.67}& \textbf{30.58} &\textbf{8.32} &\textbf{29.90} \\
SR w/o incremental refinement & 42.72& 29.00& 6.92& 28.59 \\
SR w/o reference point adjustment& 42.61& 28.62& 6.28& 28.05\\
\bottomrule
\end{tabular}
\end{table}

To investigate how the segment refinement mechanism improves the performance, we evaluate the effect of incremental refinement and reference point adjustment. As can be observed in Table~\ref{tab:segment_refine_ablation}, disabling incremental refinement leads to a 1.31\% decrease in mAP, which suggests that the incremental refinement strategy is better than independent prediction at each decoder layer. Disabling reference point adjustment in standard segment refinement mechanism also leads to a 1.85\% decrease in mAP.
This is reasonable, as the initial referent points associated with the queries are invariant for different input samples. The adjustment strategy makes them adaptive to the input and aligned with the local features of the target actions.

Besides the detection performance, another important aspect is the computation cost. In terms of FLOPs, adding or removing the two components has little impact, which is shown in Table~\ref{tab:hacs_ablation}. 
In terms of run time, the average time cost per video on THUMOS14 is 130 ms for TadTR-lite without the two components. Adding the actionness regression head will increase the average time cost to 141 ms. With segment refinement enabled, the average time cost per video becomes 155 ms, which is still very efficient compared with state-of-the-art methods. 

\begin{table}[tb]
\caption{Effects of the numbers of encoder layers and decoder layers on HACS Segments.}
\label{tab:layers_effect}
\centering
\begin{tabular}{*{2}{p{0.5cm}<{\centering}}|*{4}{p{0.8cm}<{\centering}}|c}
\toprule
$L_E$ & $L_D$  &mAP$_{0.5}$&mAP$_{0.75}$ &mAP$_{0.95}$ &mAP&MFLOPs \\

\midrule
2 & 4 & \textbf{45.16}& \textbf{30.70}& \textbf{11.78}& \textbf{30.83} &\textbf{100.5}\\
4 & 4 & 44.63& 30.39& 10.76& 30.39 &105.6\\
6 & 4 & 40.55& 27.55& 9.88 & 27.63 &110.8\\
\midrule
2 & 2 &   42.10 &29.05& 9.57& 28.84&\textbf{79.6} \\
2 & 4 & 45.16& 30.70& \textbf{11.78}& \textbf{30.83}&100.5 \\
2 & 6 &   \textbf{45.20} &\textbf{30.82} &10.67 &30.74&121.3 \\
\bottomrule
        
\end{tabular}
\end{table}

\vspace{1ex}\noindent\textbf{The effects of the number of action queries.} Table~\ref{tab:nq} compares the performance of TadTR using different number of action queries ($N_q$). The best performance is achieved at $N_q=40$ on THUMOS14, $N_q=30$ on HACS Segments and $N_q=10$ on ActivityNet-1.3. The results are reasonable, as the average number of action instances per video on THUMOS14 (15.4) is larger than that on ActivityNet-1.3 (1.5) and HACS Segments (2.8).

\begin{table}[tb]
\caption{Effect of the number of action queries on different datasets. The average mAPs are reported.}
\label{tab:nq}
\centering
\begin{tabular}{l|*{5}{p{0.85cm}<{\centering}}}
\toprule
\#queries $N_q$& 10 & 20& 30& 40&50\\
\midrule
THUMOS14&  44.06&46.48&46.94&\textbf{47.92}&46.78\\
HACS Segments& 29.63&30.73&\textbf{30.83}&29.99&29.47 \\
ActivityNet-1.3 & \textbf{28.21}&26.47&26.27&26.29&22.96\\
\bottomrule
\end{tabular}
\end{table}

\vspace{1ex}\noindent\textbf{The effects of the numbers of encoder layers and decoder layers.}
We evaluate TadTR with different numbers of encoder layers ($L_E$) and decoder layers ($L_D$) and report results in Table~\ref{tab:layers_effect}. 
With $L_D$ fixed, the best performance is achieved when $L_E$ is 2. Larger $L_E$ gives inferior results probably due to the difficulty of training. Therefore, we set $L_E$ to 2.
With $L_E$ fixed, the average mAP increases by 2.03\% when $L_D$ increases from 2 to 4. Larger $L_D$ gives a slightly lower performance. Therefore we suggest setting $L_D$ to 4.

\begin{table}[tb]
\caption{Effect of the number of sampling points $K$ on the performance and computation cost on HACS Segments. The models are trained for half of the full training cycle (15 epochs).}
\label{tab:points}
\centering
\begin{tabular}{c|*{4}{p{0.9cm}<{\centering}}|c}
\toprule
\#points $K$&mAP$_{0.5}$&mAP$_{0.75}$ &mAP$_{0.95}$ &mAP & MFLOPs\\
\midrule

1&39.22& 27.07&  9.80& 27.03&\textbf{99.1}\\
2&40.22& 27.62& 10.15& 27.61&99.6\\
4&\textbf{41.20}& \textbf{28.52}& \textbf{10.63}& \textbf{28.49}&100.5\\
8&39.77& 27.15&  9.86& 27.20&102.4\\
\bottomrule
\end{tabular}
    
\end{table}

\vspace{1ex}\noindent\textbf{The effect of the number of sampling points.}
Table~\ref{tab:points} compares the performance and computation cost using different numbers of sampling points $K$ in TDA modules on HACS Segments. We observe that moderately increasing $K$ improves the performance, as more temporal details can be captured. The best performance is achieved at $K=4$. The number of sampling points has little impact on the computation cost.

\begin{table}[tb]
\caption{Effect of the number of attention heads $M$ on the performance and computation cost on HACS Segments. The models are trained for half of the full training cycle (15 epochs).}
\label{tab:head}
\centering
\begin{tabular}{c|*{4}{p{0.9cm}<{\centering}}|c}
\toprule

\#heads $M$&mAP$_{0.5}$&mAP$_{0.75}$ &mAP$_{0.95}$ &mAP & MFLOPs \\
\midrule
1&38.62& 25.25& 7.94 & 25.57&\textbf{100.2}\\
2&40.30& 26.37& 8.09 & 26.52&100.2\\
4&40.74& 28.41& 10.51& 28.30&100.3\\
8&\textbf{41.20}& \textbf{28.52}& \textbf{10.63}& \textbf{28.49}&100.5\\
16&41.07& 28.35& 10.23& 28.31&100.9\\
\bottomrule
\end{tabular}
\end{table}

\vspace{1ex}\noindent\textbf{The effect of the number of attention heads.}
Table~\ref{tab:head} compares the performance and computation cost using different number of attention heads $M$ on HACS Segments. It is observed that moderately increasing $M$ boosts the performance, as more diverse features can be learned. The performance is saturated at $M=8$. Similar to the number of sampling points, the number of attention heads has little impact on the computation cost.

\begin{table}[]
\centering
\caption{Effect of the expanding factor $\epsilon$ and the number of bins $T_R$ for actionness regression on HACS Segments.}
\label{tab:ar_param_effect}
\begin{tabular}{c|cccc}
\toprule
$\epsilon$ & 1&1.25&1.5&2 \\
\midrule
mAP & 30.27&30.40& \textbf{30.83}&30.01 \\
\midrule
$T_R$ & 8&16&32&- \\
\midrule
mAP &30.39&\textbf{30.83}&30.26&-\\
\bottomrule
\end{tabular}
\end{table}

\vspace{1ex}\noindent\textbf{The effects of the hyper-parameters in actionness regression.} In Table~\ref{tab:ar_param_effect}, we compare different choices of the expanding factor $\epsilon$ and the number of bins $T_R$ for actionness regression. We see that the best performance is achieved when $\epsilon=1.5$. This result is 0.56\% higher than that of the variant without RoI expansion ($\epsilon=1$), showing the effectiveness of RoI expansion. Among different choices of $T_R$, $T_R=16$ results in the best performance.

\begin{figure*}
    \centering
\includegraphics[width=\linewidth]{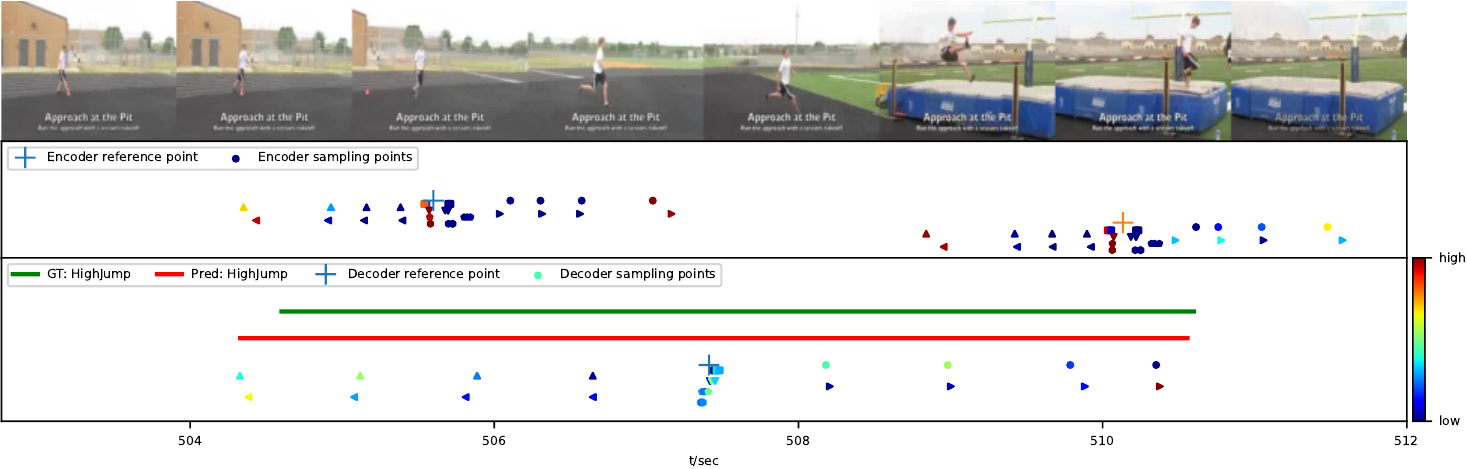}
\label{fig:attention:futsal}
\caption{Visualization of temporal deformable attention. The first row is uniformly sampled video frames. The second row visualizes the attention at two randomly picked reference points in the last encoder layer. The third row visualizes the attention for the predicted action in the last decoder layer. We use different markers to represent sampling points in different heads and separate the points from different heads vertically. The color of a point indicates the attention weight. Best viewed in color. More examples are given in the supplementary material.}
\label{fig:attention_vis}
\end{figure*}

\subsection{Analysis}
\label{subsec:exp:analysis}

\begin{figure}
    \centering
    \includegraphics[width=\linewidth]{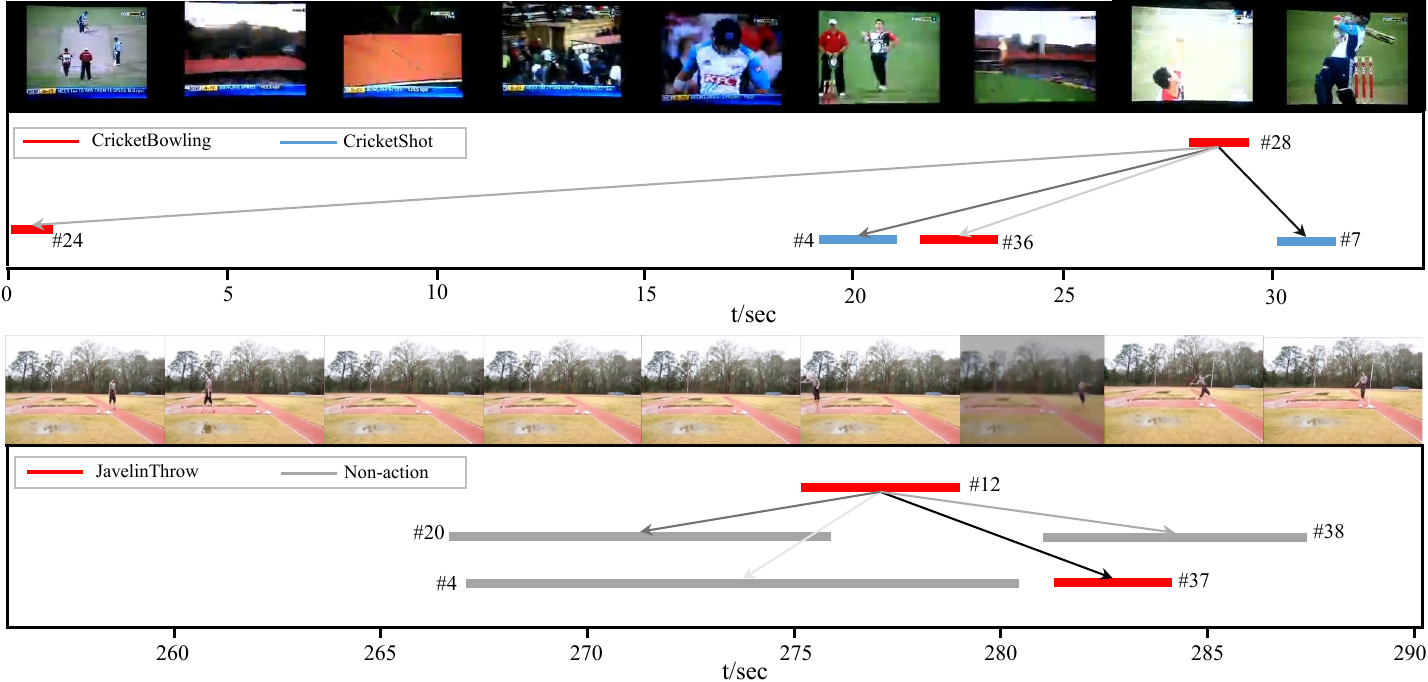}
    \caption{Visualization of self-attention between action queries in the last decoder layer. We average the attention over all heads. The queries are represented by the predicted instances. The arrows indicate the attention that the topmost query casts on the other four queries with the largest weights. The attention weight is encoded by the color. The darker the arrow, the larger the attention weight. In the first example, the four attended instances are semantically related to the topmost instance. In the second example, the topmost instance casts the most attention to a nearby instance (\#37) with the same class. The other three non-action instances are also attended, probability for context from the background. Best viewed in color.}
    \label{fig:dec_self_att}
\end{figure}

\noindent\textbf{Visualization of attention.}  Fig.~\ref{fig:attention_vis} visualizes temporal deformable attention of the last encoder layer and the last decoder layer. 
We observe that:
(1) Different attention heads focus on different temporal regions and scales. For example, the sampling points marked with left-triangle and up-triangle are distributed on the left side of the reference point. 
In some heads, the sampling points that are farthest away from the reference point have relatively higher attention weights, to capture useful cues for action boundaries.
(2) The encoder and the decoder have different preferences for context. The sampling points in the decoder almost cover the full extent of an action prediction, providing a large receptive field. Differently, the sampling points in the encoder have a relatively short temporal extent, capturing a moderate amount of context.
Fig.~\ref{fig:dec_self_att} visualizes self-attention between action queries in the last decoder layer. We find that the topmost query in each example cast the most attention on the queries whose predicted instances are semantically related to that of it. It suggests that the self-attention layer in the decoder can model the relations between action queries (or instances).

\begin{figure}[tb]
\centering
\includegraphics[width=\linewidth]{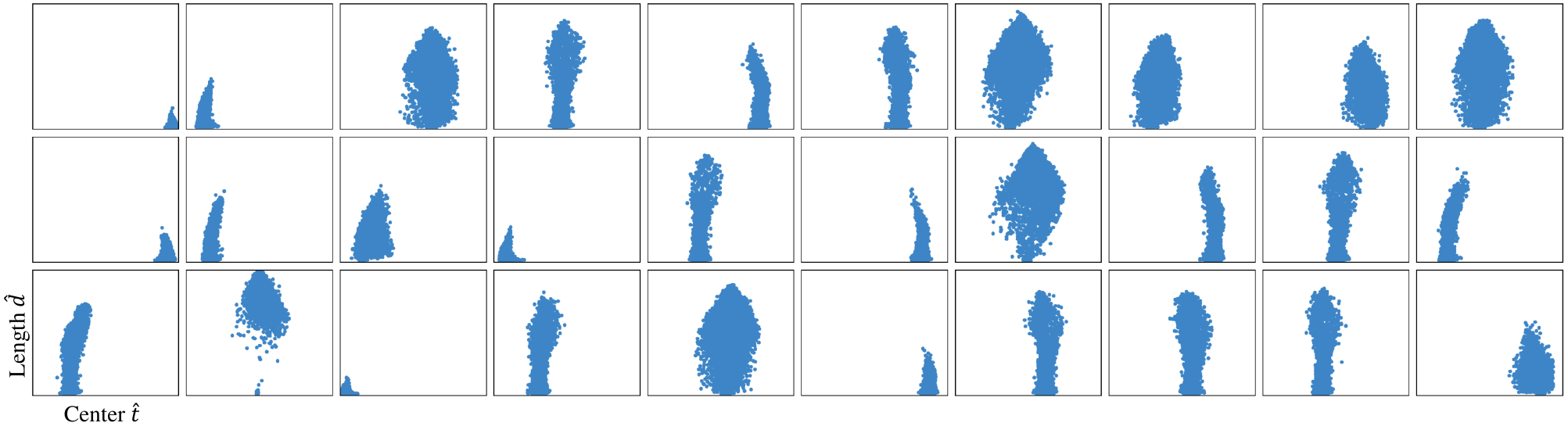}
\caption{Visualization of all action predictions on all videos from HACS Segments validation set from 30 action query slots.
In each 1-by-1 square, we use scattered points to represent all predictions from this query. The horizontal and the vertical coordinates are the coordinate of the center and the normalized length of these predictions, respectively. We observe that each query is responsible for action predictions in certain locations and lengths. 
}
\label{fig:slots}
\end{figure}

\vspace{1ex}\noindent\textbf{Visualization of action queries.} Fig.~\ref{fig:slots} illustrates the distribution of locations and scales (lengths) of output actions associated with each action query. We observe that each query produces action predictions in certain locations and scales. Different locations and scales are covered by a small number of queries. It means that the detector learns the distribution of actions in the training dataset. This is more flexible than the hand-crafted anchor design in previous methods.

\begin{figure*}[t]
\centering
\includegraphics[width=\linewidth]{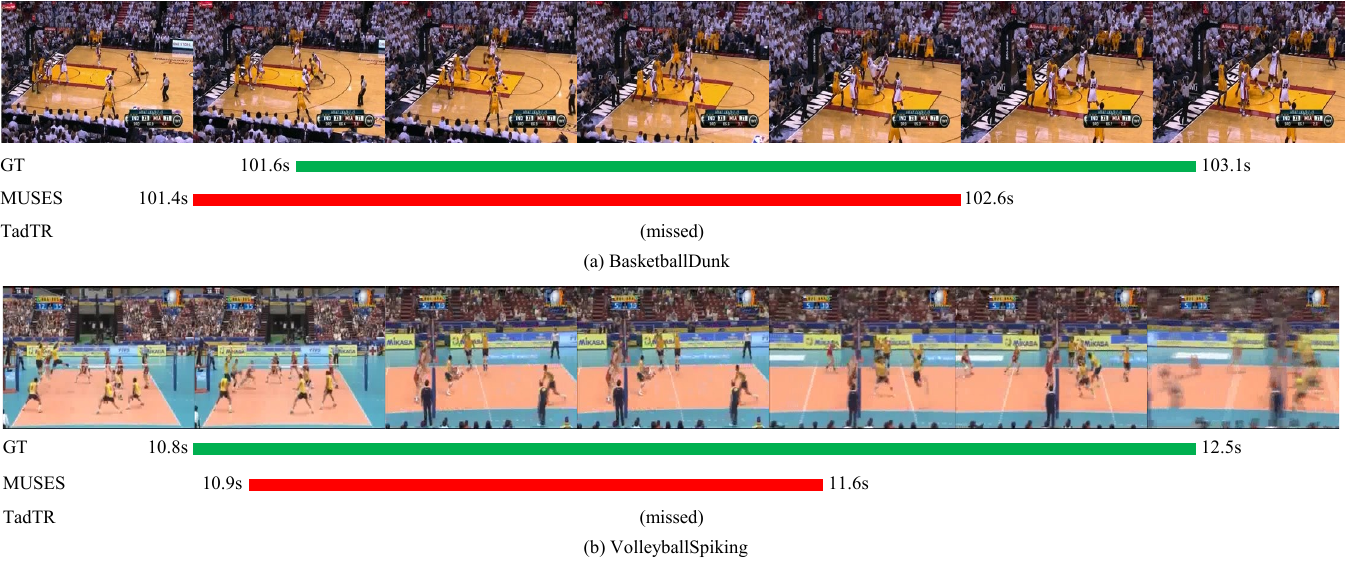}
\caption{Failure cases. TadTR misses the two short actions that are hard to detect while the 1D CNN-based method MUSES partially detects them.
}
\label{fig:failure_cases}
\end{figure*}

\vspace{1ex}\noindent\textbf{Limitations.} 
Although TadTR achieves strong overall performance, it may fail on some short actions, as depicted in Fig.~\ref{fig:failure_cases}. The quantitative results in Fig.~\ref{fig:ap_by_length} also illustrate the lower performance of TadTR on short actions. One possible reason is that Transformer is inferior to 1D CNN in modeling short-term dependency. Combing Transformer and 1D CNN might improve the performance on short actions. Another limitation is that TadTR will miss actions when the number of true  actions in a video is larger than the number of queries $N_q$, although such cases might be rare. This is a common issue of DETR-alike detectors. How to maintain the performance while increasing $N_q$ is worth studying in future works.

\section{Conclusion}
We propose TadTR, a simple end-to-end method for temporal action detection (TAD) based on Transformer. It views the TAD task as a direct set prediction problem and maps a series of learnable embeddings to action instances in parallel by adaptively extracting temporal context in the video.
It simplifies the pipeline of TAD and removes hand-crafted components such as anchor setting and post-processing. We make three improvements to enhance the Transformer with locality awareness to better adapt to the TAD task.
Extensive experiments validate the remarkable performance and efficiency of TadTR and the effectiveness of different components. TadTR achieves state-of-the-art or competitive performance on HACS Segments, THUMOS14, and ActivityNet-1.3 with lower computation costs. 
We hope that this work could trigger the development of Transformers and efficient models for temporal action detection.
The current implementation of TadTR is based on offline extracted CNN features for a fair comparison with previous methods. In the future, we plan to explore joint learning of the video encoder and TadTR~\cite{Liu_2022_CVPR}, and temporal action detectors purely based on Transformers.   

\begin{figure*}[!h]
    \centering
    \includegraphics[width=\linewidth]{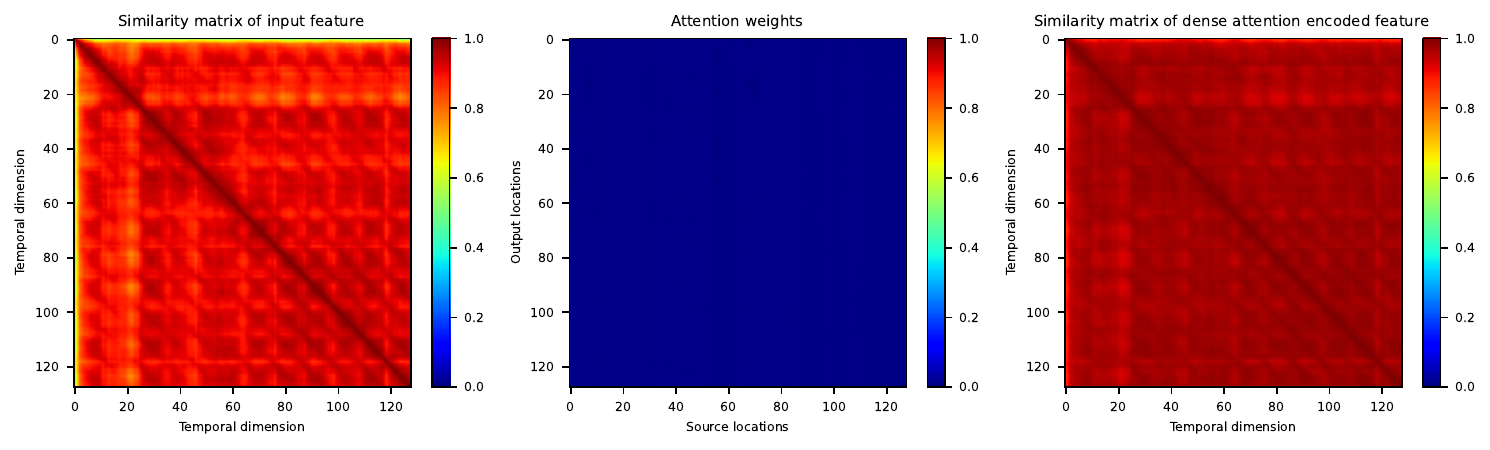}
    \caption{Different frames in a video is usually highly similar. The dense attention tends to cast uniform attention to different locations in the input sequence at initialization. Left: The similarity matrix of each pair of snippets in CNN features of a randomly selected video. Middle: The attention weight. Right: The similarity matrix of the output feature of the dense-attention. Best viewed in color.}
    \label{fig:smoothing_effect}
\end{figure*}

\appendix
In this supplement, we present several visualization results. Fig.~\ref{fig:smoothing_effect} illustrates the smoothing effect of dense attention. Fig.~\ref{fig:attention_vis} supplements Fig. 7 in the main document and gives more examples to demonstrate temporal deformable attention.

\begin{figure*}[!b]
\centering
\subfloat[Futsal]{
\includegraphics[width=\textwidth]{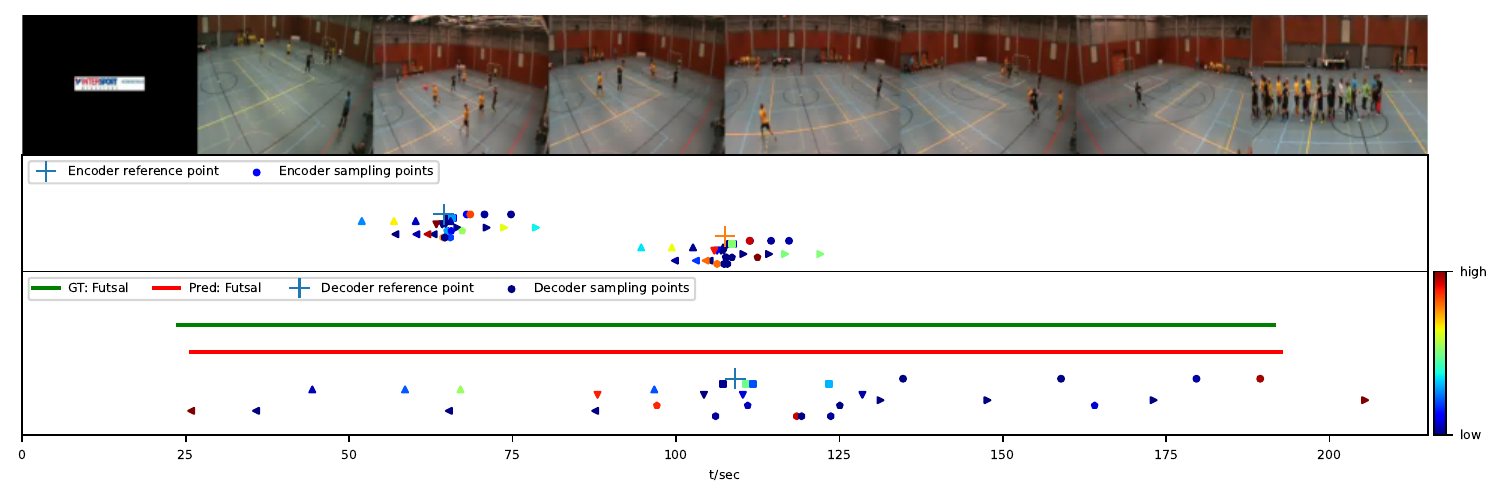}}

\subfloat[Diving]{
\includegraphics[width=\textwidth]{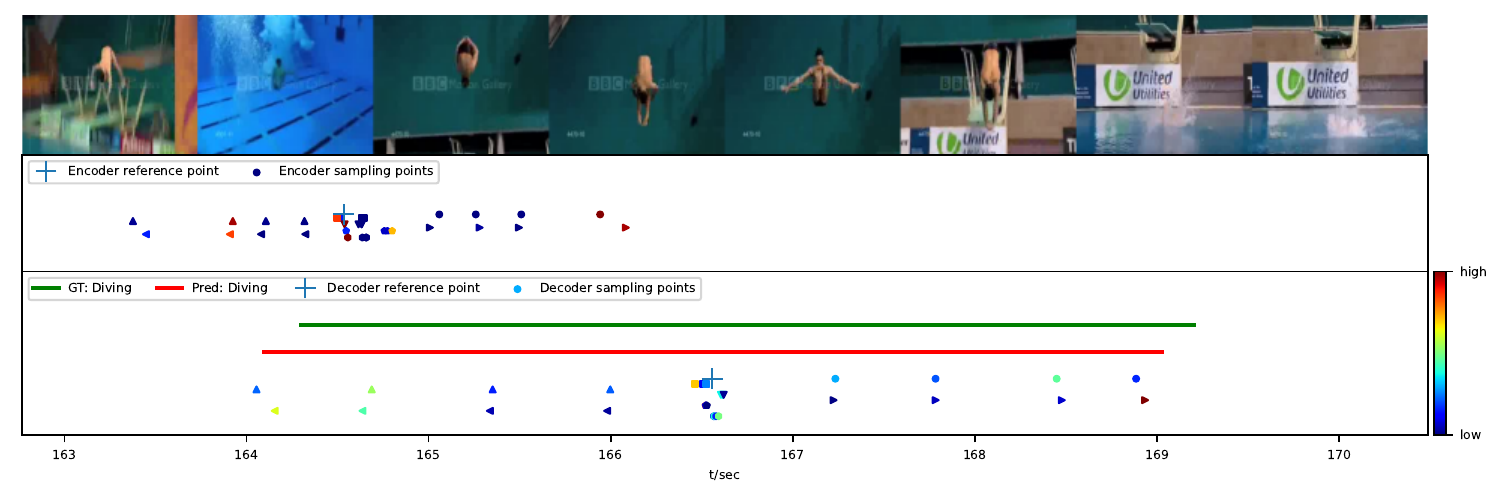}}
\caption{Visualization of temporal deformable attention. The first row is uniformly sampled video frames. The second row visualizes the attention at two randomly picked reference points in the last encoder layer. The third row visualizes the attention for the predicted action in the last decoder layer. We use different markers to represent sampling points in different heads and separate points from different heads vertically. The color of a point indicates the attention weight. Best viewed in color.}
\label{fig:attention_vis}
\end{figure*}

\bibliographystyle{IEEEtran}
\bibliography{ref}

\begin{thebibliography}{10}
\providecommand{\url}[1]{#1}
\csname url@samestyle\endcsname
\providecommand{\newblock}{\relax}
\providecommand{\bibinfo}[2]{#2}
\providecommand{\BIBentrySTDinterwordspacing}{\spaceskip=0pt\relax}
\providecommand{\BIBentryALTinterwordstretchfactor}{4}
\providecommand{\BIBentryALTinterwordspacing}{\spaceskip=\fontdimen2\font plus
\BIBentryALTinterwordstretchfactor\fontdimen3\font minus
  \fontdimen4\font\relax}
\providecommand{\BIBforeignlanguage}[2]{{%
\expandafter\ifx\csname l@#1\endcsname\relax
\typeout{** WARNING: IEEEtran.bst: No hyphenation pattern has been}%
\typeout{** loaded for the language `#1'. Using the pattern for}%
\typeout{** the default language instead.}%
\else
\language=\csname l@#1\endcsname
\fi
#2}}
\providecommand{\BIBdecl}{\relax}
\BIBdecl

\bibitem{shou2016temporal}
Z.~Shou, D.~Wang, and S.-F. Chang, ``Temporal action localization in untrimmed
  videos via multi-stage cnns,'' in \emph{CVPR}, 2016, pp. 1049--1058.

\bibitem{xu2020g}
M.~Xu, C.~Zhao, D.~S. Rojas, A.~Thabet, and B.~Ghanem, ``{G-TAD}: Sub-graph
  localization for temporal action detection,'' in \emph{CVPR}, 2020, pp.
  10\,156--10\,165.

\bibitem{Ma_2016_CVPR}
S.~Ma, L.~Sigal, and S.~Sclaroff, ``Learning activity progression in lstms for
  activity detection and early detection,'' in \emph{CVPR}, June 2016, pp.
  1942--1950.

\bibitem{richard2016temporal}
A.~Richard and J.~Gall, ``Temporal action detection using a statistical
  language model,'' in \emph{CVPR}, 2016, pp. 3131--3140.

\bibitem{caba2016fast}
F.~Caba~Heilbron, J.~Carlos~Niebles, and B.~Ghanem, ``Fast temporal activity
  proposals for efficient detection of human actions in untrimmed videos,'' in
  \emph{CVPR}, 2016, pp. 1914--1923.

\bibitem{xu2017r}
H.~Xu, A.~Das, and K.~Saenko, ``R-c3d: region convolutional 3d network for
  temporal activity detection,'' in \emph{ICCV}, 2017, pp. 5794--5803.

\bibitem{chao2018rethinking}
Y.-W. Chao, S.~Vijayanarasimhan, B.~Seybold, D.~A. Ross, J.~Deng, and
  R.~Sukthankar, ``Rethinking the faster r-cnn architecture for temporal action
  localization,'' in \emph{CVPR}, 2018, pp. 1130--1139.

\bibitem{lin2017single}
T.~Lin, X.~Zhao, and Z.~Shou, ``Single shot temporal action detection,'' in
  \emph{ACM MM}, 2017, pp. 988--996.

\bibitem{yuan2017temporal}
Z.-H. Yuan, J.~C. Stroud, T.~Lu, and J.~Deng, ``Temporal action localization by
  structured maximal sums,'' in \emph{CVPR}, 2017, pp. 3684--3692.

\bibitem{zhao2017temporal}
Y.~Zhao, Y.~Xiong, L.~Wang, Z.~Wu, X.~Tang, and D.~Lin, ``Temporal action
  detection with structured segment networks,'' \emph{ICCV}, pp. 2914--2923,
  2017.

\bibitem{lin2019bmn}
T.~Lin, X.~Liu, X.~Li, E.~Ding, and S.~Wen, ``Bmn: Boundary-matching network
  for temporal action proposal generation,'' in \emph{ICCV}, 2019, pp.
  3889--3898.

\bibitem{lea2017temporal}
C.~Lea, M.~D. Flynn, R.~Vidal, A.~Reiter, and G.~D. Hager, ``Temporal
  convolutional networks for action segmentation and detection,'' in
  \emph{CVPR}, 2017, pp. 156--165.

\bibitem{zeng2019graph}
R.~Zeng, W.~Huang, M.~Tan, Y.~Rong, P.~Zhao, J.~Huang, and C.~Gan, ``Graph
  convolutional networks for temporal action localization,'' in \emph{ICCV},
  2019, pp. 7094--7103.

\bibitem{carreira2017quo}
J.~Carreira and A.~Zisserman, ``Quo vadis, action recognition? a new model and
  the kinetics dataset,'' in \emph{CVPR}, 2017, pp. 4724--4733.

\bibitem{carion2020end}
N.~Carion, F.~Massa, G.~Synnaeve, N.~Usunier, A.~Kirillov, and S.~Zagoruyko,
  ``End-to-end object detection with transformers,'' in \emph{ECCV}, 2020, pp.
  213--229.

\bibitem{vaswani2017attention}
A.~Vaswani, N.~Shazeer, N.~Parmar, J.~Uszkoreit, L.~Jones, A.~N. Gomez,
  L.~Kaiser, and I.~Polosukhin, ``Attention is all you need,'' in \emph{NIPS},
  2017, pp. 5998--6008.

\bibitem{dai2017temporal}
X.~Dai, B.~Singh, G.~Zhang, L.~S. Davis, and Y.~Q. Chen, ``Temporal context
  network for activity localization in videos,'' in \emph{ICCV}, 2017, pp.
  5727--5736.

\bibitem{alwassel2018diagnosing}
H.~Alwassel, F.~Caba~Heilbron, V.~Escorcia, and B.~Ghanem, ``Diagnosing error
  in temporal action detectors,'' in \emph{ECCV}, 2018, pp. 256--272.

\bibitem{zhu2021deformable}
X.~Zhu, W.~Su, L.~Lu, B.~Li, X.~Wang, and J.~Dai, ``Deformable detr: Deformable
  transformers for end-to-end object detection,'' in \emph{ICLR}, 2021.

\bibitem{he2017mask}
K.~He, G.~Gkioxari, P.~Doll{\'a}r, and R.~Girshick, ``Mask r-cnn,'' in
  \emph{ICCV}, 2017, pp. 2961--2969.

\bibitem{zhao2019hacs}
H.~Zhao, A.~Torralba, L.~Torresani, and Z.~Yan, ``{HACS:} human action clips
  and segments dataset for recognition and temporal localization,'' in
  \emph{ICCV}, 2019, pp. 8667--8677.

\bibitem{jiang2014thumos}
H.~Idrees, A.~R. Zamir, Y.-G. Jiang, A.~Gorban, I.~Laptev, R.~Sukthankar, and
  M.~Shah, ``The {THUMOS} challenge on action recognition for videos “in the
  wild”,'' pp. 1--23, 2017.

\bibitem{caba2015activitynet}
F.~Caba~Heilbron, V.~Escorcia, B.~Ghanem, and J.~Carlos~Niebles,
  ``{ActivityNet}: A large-scale video benchmark for human activity
  understanding,'' in \emph{CVPR}, 2015, pp. 961--970.

\bibitem{yuan2016temporal}
J.~Yuan, B.~Ni, X.~Yang, and A.~A. Kassim, ``Temporal action localization with
  pyramid of score distribution features,'' in \emph{CVPR}, 2016, pp.
  3093--3102.

\bibitem{heilbron2017scc}
F.~C. Heilbron, W.~Barrios, V.~Escorcia, and B.~Ghanem, ``Scc: Semantic context
  cascade for efficient action detection.'' in \emph{CVPR}, 2017, pp.
  3175--3184.

\bibitem{liu2020self}
X.~Liu, Y.~Sun, J.~Lu, C.~Yao, and Y.~Zhou, ``Self-similarity action
  proposal,'' \emph{IEEE Signal Processing Letters}, vol.~27, pp. 2064--2068,
  2020.

\bibitem{Liu_2021_CVPR}
X.~Liu, Y.~Hu, S.~Bai, F.~Ding, X.~Bai, and P.~H.~S. Torr, ``Multi-shot
  temporal event localization: A benchmark,'' in \emph{CVPR}, June 2021, pp.
  12\,596--12\,606.

\bibitem{gao2017turn}
J.~Gao, Z.~Yang, C.~Sun, K.~Chen, and R.~Nevatia, ``Turn tap: Temporal unit
  regression network for temporal action proposals,'' in \emph{ICCV}, 2017, pp.
  3648--3656.

\bibitem{lin2018bsn}
T.~Lin, X.~Zhao, H.~Su, C.~Wang, and M.~Yang, ``Bsn: Boundary sensitive network
  for temporal action proposal generation,'' in \emph{ECCV}, September 2018,
  pp. 3--21.

\bibitem{gao2018ctap}
J.~Gao, K.~Chen, and R.~Nevatia, ``Ctap: Complementary temporal action proposal
  generation,'' in \emph{ECCV}, September 2018, pp. 70--85.

\bibitem{liu2019multi}
Y.~Liu, L.~Ma, Y.~Zhang, W.~Liu, and S.-F. Chang, ``Multi-granularity generator
  for temporal action proposal,'' in \emph{CVPR}, 2019, pp. 3604--3613.

\bibitem{escorcia2016daps}
V.~Escorcia, F.~C. Heilbron, J.~C. Niebles, and B.~Ghanem, ``Daps: Deep action
  proposals for action understanding,'' in \emph{ECCV}, 2016, pp. 768--784.

\bibitem{buch2017sst}
S.~Buch, V.~Escorcia, C.~Shen, B.~Ghanem, and J.~C. Niebles, ``Sst:
  Single-stream temporal action proposals,'' in \emph{CVPR}, 2017, pp.
  6373--6382.

\bibitem{long2019gaussian}
F.~Long, T.~Yao, Z.~Qiu, X.~Tian, J.~Luo, and T.~Mei, ``Gaussian temporal
  awareness networks for action localization,'' in \emph{CVPR}, 2019, pp.
  344--353.

\bibitem{lin2021learning}
C.~Lin, C.~Xu, D.~Luo, Y.~Wang, Y.~Tai, C.~Wang, J.~Li, F.~Huang, and Y.~Fu,
  ``Learning salient boundary feature for anchor-free temporal action
  localization,'' in \emph{CVPR}, 2021, pp. 3320--3329.

\bibitem{yang2020revisiting}
L.~Yang, H.~Peng, D.~Zhang, J.~Fu, and J.~Han, ``Revisiting anchor mechanisms
  for temporal action localization,'' \emph{IEEE Transactions on Image
  Processing}, vol.~29, pp. 8535--8548, 2020.

\bibitem{su2021pcg}
R.~Su, D.~Xu, L.~Sheng, and W.~Ouyang, ``Pcg-tal: Progressive cross-granularity
  cooperation for temporal action localization,'' \emph{IEEE Transactions on
  Image Processing}, vol.~30, pp. 2103--2113, 2021.

\bibitem{yeung2016end}
S.~Yeung, O.~Russakovsky, G.~Mori, and L.~Fei-Fei, ``End-to-end learning of
  action detection from frame glimpses in videos,'' in \emph{CVPR}, 2016, pp.
  2678--2687.

\bibitem{ma2020sf}
F.~Ma, L.~Zhu, Y.~Yang, S.~Zha, G.~Kundu, M.~Feiszli, and Z.~Shou, ``Sf-net:
  Single-frame supervision for temporal action localization,'' in
  \emph{ECCV}.\hskip 1em plus 0.5em minus 0.4em\relax Springer, 2020, pp.
  420--437.

\bibitem{nguyen2018weakly}
P.~Nguyen, T.~Liu, G.~Prasad, and B.~Han, ``Weakly supervised action
  localization by sparse temporal pooling network,'' in \emph{CVPR}, 2018, pp.
  6752--6761.

\bibitem{paul2018w}
S.~Paul, S.~Roy, and A.~K. Roy-Chowdhury, ``W-talc: Weakly-supervised temporal
  activity localization and classification,'' in \emph{ECCV}, September 2018,
  pp. 588--607.

\bibitem{liu2019completeness}
D.~Liu, T.~Jiang, and Y.~Wang, ``Completeness modeling and context separation
  for weakly supervised temporal action localization,'' in \emph{CVPR}, 2019,
  pp. 1298--1307.

\bibitem{shou2018autoloc}
Z.~Shou, H.~Gao, L.~Zhang, K.~Miyazawa, and S.-F. Chang, ``Autoloc:
  Weakly-supervised temporal action localization in untrimmed videos,'' in
  \emph{ECCV}, 2018, pp. 154--171.

\bibitem{yu2019temporal}
T.~Yu, Z.~Ren, Y.~Li, E.~Yan, N.~Xu, and J.~Yuan, ``Temporal structure mining
  for weakly supervised action detection,'' in \emph{ICCV}, 2019, pp.
  5522--5531.

\bibitem{huang2021modeling}
L.~Huang, Y.~Huang, W.~Ouyang, and L.~Wang, ``Modeling sub-actions for weakly
  supervised temporal action localization,'' \emph{IEEE Transactions on Image
  Processing}, vol.~30, pp. 5154--5167, 2021.

\bibitem{yang2021multi}
W.~Yang, T.~Zhang, Z.~Mao, Y.~Zhang, Q.~Tian, and F.~Wu, ``Multi-scale
  structure-aware network for weakly supervised temporal action detection,''
  \emph{IEEE Transactions on Image Processing}, vol.~30, pp. 5848--5861, 2021.

\bibitem{zeng2019breaking}
R.~Zeng, C.~Gan, P.~Chen, W.~Huang, Q.~Wu, and M.~Tan, ``Breaking
  winner-takes-all: Iterative-winners-out networks for weakly supervised
  temporal action localization,'' \emph{IEEE Transactions on Image Processing},
  vol.~28, no.~12, pp. 5797--5808, 2019.

\bibitem{huang2020relational}
L.~Huang, Y.~Huang, W.~Ouyang, and L.~Wang, ``Relational prototypical network
  for weakly supervised temporal action localization,'' in \emph{AAAI},
  vol.~34, no.~07, 2020, pp. 11\,053--11\,060.

\bibitem{islam2021hybrid}
A.~Islam, C.~Long, and R.~Radke, ``A hybrid attention mechanism for
  weakly-supervised temporal action localization,'' in \emph{AAAI}, vol.~35,
  no.~2, 2021, pp. 1637--1645.

\bibitem{zhai2020two}
Y.~Zhai, L.~Wang, W.~Tang, Q.~Zhang, J.~Yuan, and G.~Hua, ``Two-stream
  consensus network for weakly-supervised temporal action localization,'' in
  \emph{ECCV}.\hskip 1em plus 0.5em minus 0.4em\relax Springer, 2020, pp.
  37--54.

\bibitem{huang2021foreground}
L.~Huang, L.~Wang, and H.~Li, ``Foreground-action consistency network for
  weakly supervised temporal action localization,'' in \emph{ICCV}, 2021, pp.
  8002--8011.

\bibitem{tan2021planetr}
B.~Tan, N.~Xue, S.~Bai, T.~Wu, and G.-S. Xia, ``Planetr: Structure-guided
  transformers for 3d plane recovery,'' in \emph{CVPR}, 2021, pp. 4186--4195.

\bibitem{chen2022transmix}
J.-N. Chen, S.~Sun, J.~He, P.~H. Torr, A.~Yuille, and S.~Bai, ``Transmix:
  Attend to mix for vision transformers,'' in \emph{CVPR}, 2022, pp.
  12\,135--12\,144.

\bibitem{liang2022transcrowd}
D.~Liang, X.~Chen, W.~Xu, Y.~Zhou, and X.~Bai, ``Transcrowd: weakly-supervised
  crowd counting with transformers,'' \emph{Science China Information
  Sciences}, vol.~65, no.~6, pp. 1--14, 2022.

\bibitem{sun2019videobert}
C.~Sun, A.~Myers, C.~Vondrick, K.~Murphy, and C.~Schmid, ``Videobert: A joint
  model for video and language representation learning,'' in \emph{ICCV}, 2019,
  pp. 7464--7473.

\bibitem{bertasius2021space}
G.~Bertasius, H.~Wang, and L.~Torresani, ``Is space-time attention all you need
  for video understanding?'' in \emph{ICML}, July 2021, pp. 813--824.

\bibitem{wu2021seqformer}
J.~Wu, Y.~Jiang, W.~Zhang, X.~Bai, and S.~Bai, ``Seqformer: a frustratingly
  simple model for video instance segmentation,'' \emph{arXiv preprint
  arXiv:2112.08275}, 2021.

\bibitem{zhu2020actbert}
L.~Zhu and Y.~Yang, ``Actbert: Learning global-local video-text
  representations,'' in \emph{CVPR}, 2020, pp. 8746--8755.

\bibitem{zhou2018end}
L.~Zhou, Y.~Zhou, J.~J. Corso, R.~Socher, and C.~Xiong, ``End-to-end dense
  video captioning with masked transformer,'' in \emph{CVPR}, 2018, pp.
  8739--8748.

\bibitem{girdhar2019video}
R.~Girdhar, J.~Carreira, C.~Doersch, and A.~Zisserman, ``Video action
  transformer network,'' in \emph{CVPR}, 2019, pp. 244--253.

\bibitem{nawhal2021activity}
M.~Nawhal and G.~Mori, ``Activity graph transformer for temporal action
  localization,'' \emph{arXiv preprint arXiv:2101.08540}, 2021.

\bibitem{tan2021relaxed}
J.~Tan, J.~Tang, L.~Wang, and G.~Wu, ``Relaxed transformer decoders for direct
  action proposal generation,'' in \emph{ICCV}, October 2021, pp.
  13\,526--13\,535.

\bibitem{wang2021temporal}
L.~Wang, H.~Yang, W.~Wu, H.~Yao, and H.~Huang, ``Temporal action proposal
  generation with transformers,'' \emph{arXiv preprint arXiv:2105.12043}, 2021.

\bibitem{bai2020boundary}
Y.~Bai, Y.~Wang, Y.~Tong, Y.~Yang, Q.~Liu, and J.~Liu, ``Boundary content graph
  neural network for temporal action proposal generation,'' in \emph{ECCV},
  2020, pp. 121--137.

\bibitem{ba2016layer}
J.~L. Ba, J.~R. Kiros, and G.~E. Hinton, ``Layer normalization,'' \emph{arXiv
  preprint arXiv:1607.06450}, 2016.

\bibitem{lin2017focal}
T.-Y. Lin, P.~Goyal, R.~Girshick, K.~He, and P.~Doll{\'a}r, ``Focal loss for
  dense object detection,'' in \emph{ICCV}, 2017, pp. 2980--2988.

\bibitem{wang2016temporal}
L.~Wang, Y.~Xiong, Z.~Wang, Y.~Qiao, D.~Lin, X.~Tang, and L.~Van~Gool,
  ``Temporal segment networks: Towards good practices for deep action
  recognition,'' in \emph{ECCV}, 2016, pp. 20--36.

\bibitem{loshchilov2017decoupled}
I.~Loshchilov and F.~Hutter, ``Decoupled weight decay regularization,'' in
  \emph{ICLR}, 2017, pp. 1--18.

\bibitem{zhao2017cuhk}
Y.~Zhao, B.~Zhang, Z.~Wu, S.~Yang, L.~Zhou, S.~Yan, L.~Wang, Y.~Xiong, W.~Yali,
  D.~Lin, Y.~Qiao, and X.~Tang, ``{CUHK} \& {ETHZ} \& {SIAT} submission to
  {ActivityNet} challenge 2017,'' \emph{arXiv preprint arXiv:1710.08011}, pp.
  20--24, 2017.

\bibitem{zhao2020bottom}
P.~Zhao, L.~Xie, C.~Ju, Y.~Zhang, Y.~Wang, and Q.~Tian, ``Bottom-up temporal
  action localization with mutual regularization,'' in \emph{ECCV}, 2020.

\bibitem{alwassel_2021_tsp}
H.~Alwassel, S.~Giancola, and B.~Ghanem, ``Tsp: Temporally-sensitive
  pretraining of video encoders for localization tasks,'' in \emph{ICCV
  Workshops}, 2021, pp. 3166--3176.

\bibitem{shou2017cdc}
Z.~Shou, J.~Chan, A.~Zareian, K.~Miyazawa, and S.-F. Chang, ``Cdc:
  Convolutional-de-convolutional networks for precise temporal action
  localization in untrimmed videos,'' in \emph{ICCV}, 2017, pp. 1417--1426.

\bibitem{Liu_2022_CVPR}
X.~Liu, S.~Bai, and X.~Bai, ``An empirical study of end-to-end temporal action
  detection,'' in \emph{CVPR}, June 2022, pp. 20\,010--20\,019.

\end{thebibliography}

\vfill
\end{document}